\pdfoutput=1
\documentclass[acmtog]{acmart}

\usepackage{booktabs} % For formal tables

\acmPrice{15.00}

\setcopyright{rightsretained} 
\acmJournal{TOG}
\acmYear{2018}
\acmVolume{37}
\acmNumber{6}
\acmArticle{255}
\acmMonth{11}
\acmDOI{10.1145/3272127.3275099}

\setcitestyle{square}

\newcommand{\etal}{et al. }
\usepackage{hyperref}

\makeatletter

\begin{document}

% Title. 
% If your title is long, consider \title[short title]{full title} - "short title" will be used for running heads.
\title{\textit{LookinGood}: Enhancing Performance Capture with Real-time Neural Re-Rendering}

%\author{Anonymous Submission}
\author{Ricardo Martin-Brualla}
\authornote{Authors equally contributed to this work.}
\author{Rohit Pandey}
\authornotemark[1]
\author{Shuoran Yang}
\author{Pavel Pidlypenskyi}
\author{Jonathan Taylor}
\author{Julien Valentin}
\author{Sameh Khamis}
\author{Philip Davidson}
\author{Anastasia Tkach}
\author{Peter Lincoln}
\author{Adarsh Kowdle}
\author{Christoph Rhemann}
\author{Dan B Goldman}
\author{Cem Keskin}
\author{Steve Seitz}
\author{Shahram Izadi}
\author{Sean Fanello}
\affiliation{\institution{Google Inc.}}
%\authorsaddresses{}
\renewcommand{\shortauthors}{Martin-Brualla, Pandey et al.}



% abstract
\begin{abstract}
Motivated by augmented and virtual reality applications such as telepresence, there has been a recent focus in \emph{real-time} performance capture of humans under motion. However, given the real-time constraint, these systems often suffer from artifacts in geometry and texture such as holes and noise in the final rendering, poor lighting, and low-resolution textures. We take the novel approach to augment such real-time performance capture systems with a deep architecture that takes a rendering from an arbitrary viewpoint, and jointly performs completion, super resolution, and denoising of the imagery in real-time. We call this approach \emph{neural (re-)rendering}, and our live system ``LookinGood". 
Our deep architecture is trained to produce high resolution and high quality images from a coarse rendering in real-time. First, we propose a self-supervised training method that does not require manual ground-truth annotation. We contribute a specialized reconstruction error that uses semantic information to focus on relevant parts of the subject, e.g. the face. We also introduce a \textit{salient reweighing scheme} of the loss function that is able to discard outliers. We specifically design the system for virtual and augmented reality headsets where the consistency between the left and right eye plays a crucial role in the final user experience. Finally, we generate temporally stable results by explicitly minimizing the difference between two consecutive frames.
We tested the proposed system in two different scenarios: one involving a single RGB-D sensor, and upper body reconstruction of an actor, the second consisting of full body $360^\circ$ capture.
Through extensive experimentation, we demonstrate how our system generalizes across unseen sequences and subjects. The supplementary video is available at \href{http://youtu.be/Md3tdAKoLGU}{\texttt{http://youtu.be/Md3tdAKoLGU}}.
\end{abstract}

%keywords

% A "teaser" figure, centered below the title and authors and above the body of the work.
\begin{teaserfigure}
  \centering
  \includegraphics[width=0.87\linewidth]{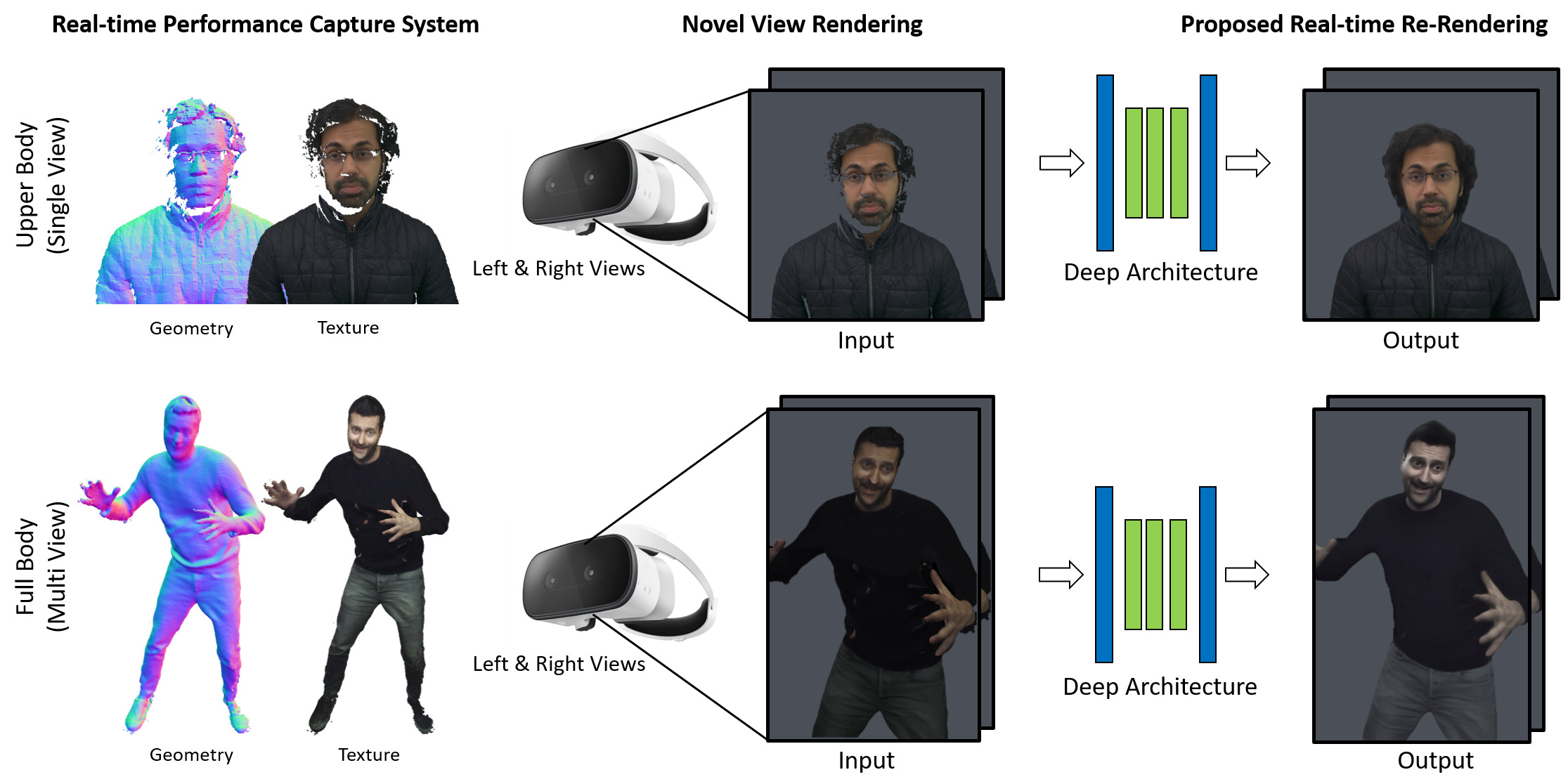}
  \caption{
  \textit{LookinGood} leverages recent advances in real-time 3D performance capture and machine learning to re-render high quality novel viewpoints of a captured scene. A textured 3D reconstruction is first rendered to a novel viewpoint. Due to imperfections in geometry and low-resolution texture, the 2D rendered image contains artifacts and is low quality. Therefore we propose a deep learning technique that takes these images as input and generates more visually enhanced re-rendering. The system is specifically designed for VR and AR headsets, and accounts for consistency between two stereo views.}
  \label{fig:teaser}
\end{teaserfigure}

% Processes all of the front-end information and starts the body of the work.
\maketitle

% !TEX root = main.tex
\section{Introduction}
\begin{figure}[t]
\centering
\includegraphics[width=\columnwidth]{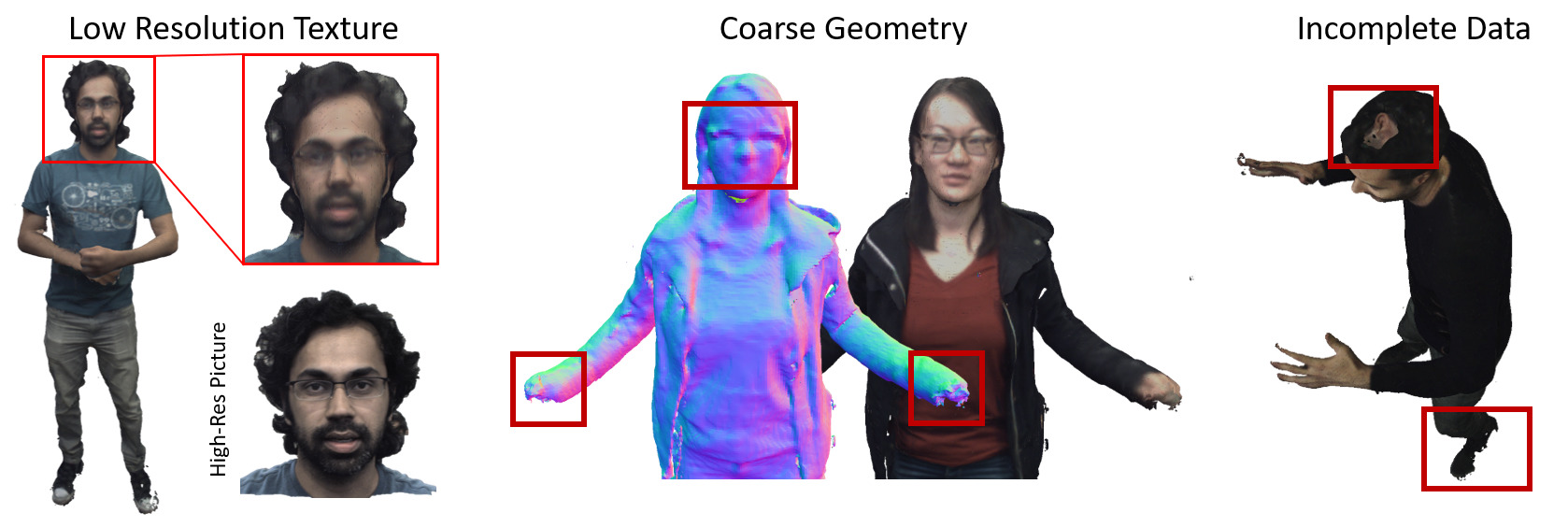}
\caption{
Limitations of state of the art, real-time performance capture systems. Left: low resolution textures where the final rendering does not resemble a high quality picture of the subject. Middle: coarse geometry leads to overly smooth surfaces where important details such as glasses are lost. This also limits the quality of the final texture. Right: incomplete data in the reconstruction creates holes in the final output.}
\label{fig:problems}
\end{figure}

The rise of augmented reality (AR) and virtual reality (VR) has created a demand for high quality 3D content of humans using performance capture rigs.   
There is a large body of work on offline multiview performance capture systems~\cite{fvv, prada, Carranza2003FreeViewpointVideo}. However, recently, real-time performance capture systems~\cite{dou16,holoportation,dou17,zollhoefer2014,dynamicfus} have opened-up new use cases for telepresence ~\cite{holoportation}, augmented videos~\cite{face2face,synthesizingobama} and live performance broadcasting~\cite{IntelFreeD}. Despite all of these efforts, the results of performance capture systems still suffer from some combination of distorted geometry~\cite{holoportation}, poor texturing and inaccurate lighting, making it difficult to reach the level of quality required in AR and VR applications.  Ultimately, this affects the final user experience (see Fig. \ref{fig:problems}). 

%Standard computer graphics techniques allow artists to create high quality synthetic objects and coarse, but evocative avatars.
%These can then be placed in virtual environments, possibly even interacting with the real world.

%However, when it comes to producing 3D models of more organic objects, with less structured shapes such as humans performing arbitrary dynamic actions, current scanning technologies that are setting the state of the art can produce results that might be uncanny or contain noticeable and distracting artifacts.  

An alternative approach consists of using controlled lighting capture stages.
The incredible results these systems produce have often been used in Hollywood productions~\cite{lightstage1,lightstage2}. However these systems are not suitable for real-time scenarios and often the underlying generated geometry is only a rough proxy, rather than an accurate reconstruction. This makes the methods difficult to apply to AR and VR scenarios where geometry and scale play a crucial role.

In this paper, we explore a hybrid direction that first leverages recent advances in real-time performance capture to obtain approximate geometry and texture in real time -- acknowledging that the final 2D rendered output of such systems will be low quality due to geometric artifacts, poor texturing and inaccurate lighting. We then leverage recent advances in deep learning to ``enhance" the final rendering to achieve higher quality results in real-time.  In particular, we use a deep architecture that takes as input the final 2D rendered image from a single or multiview performance capture system, and learns to enhance such imagery in real-time, producing a final high quality re-rendering (see Fig. \ref{fig:teaser}). We call this approach \emph{neural re-rendering}, and we demonstrate state of the art results within two real-time performance capture systems -- one single RGB-D and one multiview.  

In summary the paper makes the following contributions:
\begin{itemize}
    \item A novel approach called neural re-rendering that learns to enhance low-quality output from performance capture systems in real-time, where images contain holes, noise, low resolution textures, and color artifacts. As a byproduct we also predict a binary segmentation mask at test-time that isolates the user from the rest of the background.
    \item A method for reducing the overall bandwidth and computation required of such a deep architecture, by forcing the network to learn the mapping from low-resolution input images to high-resolution output renderings. At test time, however, only the low-resolution images are used from the live performance capture system.
    \item A specialized loss function that uses semantic information to produce high quality results on faces. To reduce the effect of outliers we propose a saliency reweighing scheme that focuses the loss on the most relevant regions.
    \item A specialized design for VR and AR headsets, where the goal is to predict two consistent views of the same object.
    \item Temporally stable re-rendering by enforcing consistency between consecutive reconstructed frames.
    \item Exhaustive experiments using two different real-time capture systems: one involving a full 360 multi-view reconstruction of the full body, and another using a single RGB-D sensor for upper body reconstructions.
\end{itemize}

% !TEX root = main.tex
\section{Related Work}
Generating high quality output from textured 3D models is the ultimate goal of many performance capture systems. Here we briefly review methods as follows: image-based approaches, full 3D reconstruction systems and finally learning based solutions.

% \paragraph{Image Based Rendering.} The seminal systems proposed in \cite{debevec1,debevec2} paved the way to many followup research trends on the topic, which is still active and very challenging. These methods, find their culmination in very sophisticated systems such as the Lightstage proposed in \cite{lightstage1}. Although capable of generating very high quality re-rendered images, these systems usually require multiple shots to infer detailed surface normals and reflective properties \cite{lightstage1}, or they use specialized hardware \cite{lightstage2} and ultimately they do not generate a consistent and accurate geometry of the objects of interested. Due to the proxy geometry, these method cannot be directly employed in AR and VR scenarios.

% Another trend solves the texturing of an object with known geometry using a Conditional Random Field (CRF) model \cite{seamless,texturemontage}. In \cite{seamless} they rely on a projective texturing approach, where multiple images are blent together according to a certain energy function, whereas in  \cite{texturemontage} authors use a sparse set of correspondences. Despite the impressive results, they require a non negligible amount of resources and multiple seconds (or minutes) per frame.

% Optical flow based methods \cite{eisemann,casas,volino}, are a viable alternative, however their accuracy is usually limited by the optical flow quality. Moreover these algorithms are restricted to off-line applications.

\paragraph{Image-based Rendering (IBR)} IBR techniques~\cite{DebevecFacade,GortlerLumigraph} warp a series of input color images to novel viewpoints of a scene using geometry as a proxy. Zitnick~\etal~\shortcite{zitnick04} expanded these methods to video inputs, where a performance is captured with multiple RGB cameras and proxy depth maps are estimated for every frame in the sequence. This work is limited to a small $30^\circ$ coverage, and its quality strongly degrades when the interpolated view is far from the original cameras. 

More recent works~\cite{eisemann,casas,volino} introduced optical flow methods to IBR, however their accuracy is usually limited by the optical flow quality. Moreover these algorithms are restricted to off-line applications.

Another limitation of IBR techniques is their use of all input images in the rendering stage, making them ill-suited for real-time VR or AR applications as they require transferring all camera streams, together with the proxy geometry. However, IBR techniques have been successfully applied to constrained applications like $360^\circ$ degree stereo video~\cite{anderson2016jump,megastereo}, which produce two separate video panoramas --- one for each eye --- but are constrained to a single viewpoint.

% \paragraph{Volumetric Videos.} Volumetric systems usually use a multiview setup that comprehends multiple RGB cameras, sometimes coupled with depth sensors. An early example of real-time re-rendering system that explicitly makes use of geometry can be found in \cite{zitnick04}. This work is limited to a small $30\deg$ coverage, and its quality strongly degrades when the interpolated view is far from the original cameras. 

\paragraph{Volumetric Capture} 
Two recent works from Microsoft~\cite{fvv,prada} use more than 100 cameras to generate high quality offline volumetric performance capture. Collet~\etal~\shortcite{fvv} used a controlled environment with green screen and carefully adjusted lighting conditions to produce high quality renderings. Their method produces rough point clouds via multi-view stereo, that is then converted into a mesh using Poisson Surface Reconstruction~\cite{poisson_surface_reconstruction}. Based on the current topology of the mesh, a keyframe is selected which is tracked over time to mitigate inconsistencies between frames. The overall processing time is $\sim 28$ minutes per frame. Prada~\etal~\shortcite{prada} extended the previous work to support texture tracking. These frameworks then deliver high quality volumetric captures at the cost of sacrificing real-time capability.

Recent proposed methods deliver performance capture in real-time~\cite{zollhoefer2014,dynamicfus,dou16,dou17,holoportation,Du2018Montage4D}. Several use single RGB-D sensors to either track a template mesh or reference volume~\cite{zollhoefer2014,dynamicfus,volumedeform,kaiwen17}. However, these systems require careful motions and none support high quality texture reconstruction. The systems of Dou~\etal\shortcite{dou16} and Orts-Escolano~\etal~\shortcite{holoportation} use fast correspondence tracking \cite{patchCollider} to extend the single view non-rigid tracking pipeline proposed by Newcombe~\etal~\shortcite{dynamicfus} to handle topology changes robustly. This method however, suffers from both geometric and texture inconsistency, as demonstrated by Dou~\etal\shortcite{dou17} and Du~\etal~\shortcite{Du2018Montage4D}. 

Even in the latest state of the art work of Dou~\etal~\shortcite{dou17} the reconstruction suffers from geometric holes, noise, and low quality textures. Du~\etal~\shortcite{Du2018Montage4D} extend previous work and propose a real-time texturing method that can be applied on top of the volumetric reconstruction to improve quality further. This is based on a simple Poisson blending scheme, as opposed to offline systems that use a Conditional Random Field (CRF) model~\cite{seamless,texturemontage}. The final results are still coarse in terms of texture. Moreover these algorithms require streaming all of the raw input images, which means it does not scale with high resolution input images.

\paragraph{Learning Based Methods}
Learning-based solutions to generate high quality renderings have shown very promising results since the groundbreaking work of Dosovitskiy~\etal\shortcite{learningchairs}.  That work, however, models only a few, explicit object classes, and the final results do not necessary resemble high-quality real objects. Followup work~\cite{kulkarni15,yang15,tatarchenko16} use end-to-end encoder-decoder networks to generate novel views of an image starting from a single viewpoint. However, due to the large variability, the results are usually low resolution. 

More recent work~\cite{deepiewmorphing,transfgrounded,appearanceflow} employ some notion of 3D geometry in the end-to-end process to deal with the 2D-3D object mapping. For instance, Zhou~\etal\shortcite{appearanceflow} use an explicit flow that maps pixels from the input image to the output novel view. In Deep View Morphing~\cite{deepiewmorphing} two input images and an explicit rectification stage, that roughly aligns the inputs, are used to generate intermediate views. Park~\etal\shortcite{transfgrounded} split the problem between visible pixels, i.e. those that can be explicitly copied from the input image, and occluded regions, i.e. areas that need to be inpainted. Another trend explicitly employs multiview stereo in an end-to-end fashion to generate intermediate view of city landscapes~\cite{deepstereo}.

3D shape completion methods~\cite{HighResShapeCompletion, angela_dai,octnet} use 3D filters to volumetrically complete 3D shapes. But given the cost of such filters both at training and at test time, these have shown low resolution reconstructions and performance far from real-time. PointProNets~\cite{PointProNets} show impressive results for denoising point clouds but again are computationally demanding, and do not consider the problem of texture reconstruction. 

The problem we consider is also closely related to the image-to-image translation task~\cite{pix2pix2016,ganvsperc,CycleGAN2017}, where the goal is to start from input images from a certain domain and ``translate" them into another domain, e.g. from semantic segmentation labels to realistic images. Our scenario is similar, as we transform low quality 3D renderings into higher quality images.

Despite the huge amount of work on the topic, it is still challenging to generate high quality renderings of people in real-time for performance capture. Contrary to previous work, we leverage recent advances in real-time volumetric capture and use these systems as input for our learning based framework to generate high quality, real-time renderings of people performing arbitrary actions.

% !TEX root = main.tex
\begin{figure}[t]
\centering
\includegraphics[width=\columnwidth]{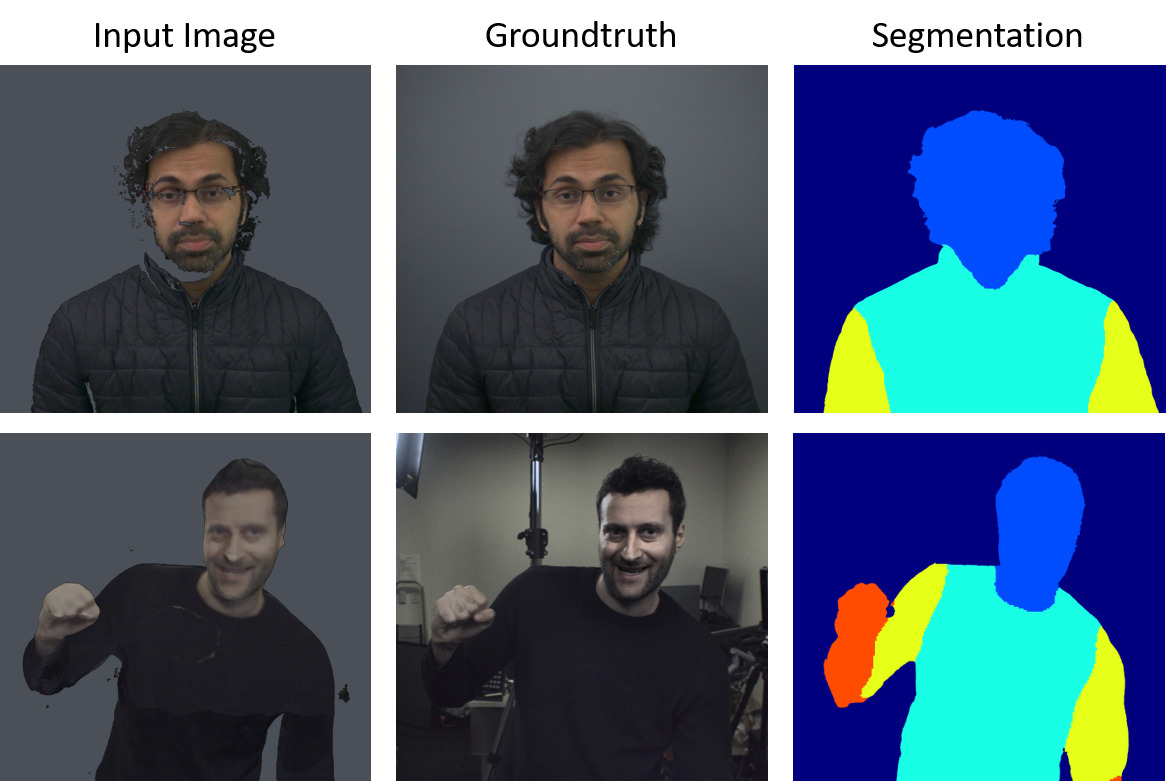}
\caption{Training data for neural re-rendering: Rendering from a volumetric reconstruction used as input, image from a witness camera used as reconstruction target, and the corresponding ground truth segmentation.}
\label{fig:training_example}
\end{figure}
\section{LookinGood with Neural Re-Rendering}
Existing real-time single and multiview performance capture pipelines \cite{dou17,dou16,holoportation,dynamicfus}, estimate the geometry and texture map of the scene being captured; this is sufficient to render that textured scene into any arbitrary (virtual) camera. Although extremely compelling, these rendering usually suffer from final artifacts, coarse geometric details, missing data, and relatively coarse textures. Examples of such problems are depicted in Fig. \ref{fig:problems}. We propose to circumvent all of these limitations using a machine learning framework called neural re-rendering. The instantiation of this machine learning based approach is a new system called \emph{LookinGood} that demonstrates unprecedented performance capture renderings in real-time. 

We focus exclusively on human performance capture and apply the proposed technique to two scenarios: a) a single RGB-D image of a person's upper body and b) another one where a person's complete body is captured by a $360^\circ$ capture setup. In the following we describe the main components of our approach.

\subsection{Learning to Enhance Reconstructions}
\label{sec:ml_prob}
In order to train our system, we placed, into the capture setup, additional ground-truth cameras that can optionally be higher resolution than the ones already in the capture rig. The proposed framework learns to map the low-quality renderings of the 3D model, captured with the rig, to a high-quality rendering at test time. 

%Volumetric capture systems output textured 3D models that can be rendered from arbitrary viewpoints, allowing a user to move freely around an object. However, current systems suffer from a variety of artifacts including lack of details, coarse geometry, incomplete data and, in general, much lower quality compared to high resolution video, as illustrated in \ref{fig:problems}. Recent advantages in Machine Learning however open new possibilities of coupling these volumetric reconstruction systems with neural networks to produce significantly higher quality results. In this paper, we propuse such a system that casts the photo-realistic reconstruction problem as a refining of the output of volumetric reconstruction systems.

The idea of casting image denoising, restoration or super-resolution to a regression task has been extensively explored in the past~\cite{super_forest,BMVC2015_58,dai_regressor,filterforest,nowozin_loss}. Compared with previous work, the problem at hand is significantly more challenging than the tasks tackled by prior art since it consists of jointly denoising, superresolving, and inpainting. Indeed, the rendered input images can be geometrically imprecise, be noisy, contain holes and be of lower resolution than the targeted output.

\subsubsection*{Witness Cameras as Groundtruth.} Ultimately, our goal is to output a high quality image in real-time given low quality input. A key insight of our approach is the use of extra cameras providing ground truth, that allow for evaluation and training of our proposed neural re-rendering task. To this end, we mount additional ``witness'' color cameras to the existing capture rigs, that capture higher quality images from different viewpoints. Note that the images captured by the witness cameras are \emph{not} used in the real-time system, and only used for training. %Indeed, online streaming of witness camera data might nor be is not feasible with the current networking technology.

\subsection{Image Enhancement}
\label{sec:imen}
\newcommand{\outputimage}{I_{e}}

Given an image $I$ rendered from a volumetric reconstruction, we want to compute an enhanced version of $I$, that we denote by $\outputimage$.

When defining the transformation function between $I$ and $\outputimage$ we specifically target VR and AR applications. We therefore define the following principles: a) the user typically focuses more on salient features, like faces, and artifacts in those areas should be highly penalized, b) when viewed in stereo, the outputs of the network have to be consistent between left and right pairs to prevent user discomfort, and c) in VR applications, the renderings are composited into the virtual world, requiring accurate segmentation masks. Finally, like in any image synthesis system, we will require our outputs to be temporally consistent.

We define the synthesizing function $F(I)$ to generate a color image $I_{pred}$ and a segmentation mask $M_{pred}$ that indicates foreground pixels such that $\outputimage = I_{pred} \odot M_{pred}$ where $\odot$ is the element-wise product, such that background pixels in $\outputimage$ are set zero. In the rest of this section, we define the training of a neural network that computes $F(I)$.

%Given our focus on VR and AR applications, we want our network to produce accurate foreground segmentations to allow the compositing of the photo-realistic volumetric video in virtual and augmented environments. To that end, 
At training time, we use a state of the art body part semantic segmentation algorithm \cite{deeplabv3} to generate $I_{seg}$, the semantic segmentation of the ground-truth image $I_{gt}$ captured by the witness camera, as illustrated in the right of Fig. \ref{fig:training_example}. To obtain improved segmentation boundaries for the subject, we refine the predictions of this algorithm using the pairwise CRF proposed by Kr\"{a}henb\"{u}hl and Koltun~\shortcite{meanfield}.

Note that at test time, this semantic segmentation is not required. However, our network does predict a binary segmentation mask as a biproduct, which can be useful for AR/VR rendering. 

%Since we are exclusively focusing on humans, we do not want the network to learn anything about the background surrounding the performer. To this end, we employ the method described in \cite{semanticseg} which performs people detection and provides labels for different body parts as illustrated in the right of Fig. \ref{fig:training_example}. In order to obtain a final segmentation that better preserves the boundaries/edges of the subject, we refine the predictions from \cite{semanticseg} using the pairwise CRF described in \cite{meanfield}.

%To train our model, we designed a loss function that satisfies the following guiding principals: 1) Photo-realistic faces are more important to the user than photo-realistc bodies. 2) The background is not important and should not affect the results. 3) Temporally stable renderings are important to the user. 4) For VR/AR applications, the stereo predictions must be consistent. 5) The loss function must be robust to outliers.

To optimize for $F(I)$, we train a neural network to optimize the loss function
\begin{align}
\begin{split}
\mathcal{L}  = & w_{1}\mathcal{L}_{rec} + w_{2} \mathcal{L}_{mask} + w_{3} \mathcal{L}_{head} + \\
 & + w_{4} \mathcal{L}_{temporal}  + w_{5} \mathcal{L}_{stereo},
\label{eq:total_loss}
\end{split}
\end{align}
where the weights $w_i$ are empirically chosen such that all the losses provide a similar contribution.

\newcommand{\ind}[1]{\mathcal{I}}
\newcommand{\VGG}{VGG}

\subsubsection*{Reconstruction Loss $\mathcal{L}_{rec}$.} Following recent advances in image reconstruction \cite{perceptual}, instead of using standard $\ell_2$ or $\ell_1$ losses in the image domain, we compute the $\ell_1$ loss in the feature space of VGG16 trained on ImageNet~\cite{imagenet_cvpr09}. Similar to related work~\cite{perceptual}, we compute the loss as the $\ell$-1 distance of the activations of conv1 through conv5 layers. This gives very comparable results to using a GAN loss~\cite{gan}, without the overhead of employing a GAN architecture during training \cite{ganvsperc}. We compute the loss as
\begin{equation}
\begin{split}
    \mathcal{L}_{rec} = \sum_{i=1}^5 \Vert \VGG_{i}( M_{gt} \odot I_{gt}) - \VGG_{i}(M_{pred} \odot I_{pred}) \Vert_{*}.
\end{split}
\end{equation}
where $M_{gt} = (I_{seg} \ne \text{background})$ is a binary segmentation mask that turns off background pixels (see Fig. \ref{fig:training_example}), $M_{pred}$ is the predicted binary segmentation mask, $\VGG_i(\cdot)$ maps an image to the activations of the conv-i layer of VGG and $\|\cdot\|_{*}$ is a ``saliency re-weighted'' $\ell_1$-norm defined later in this section. To speed-up color convergence, we optionally add a second term to $\mathcal{L}_{rec}$ defined as the $\ell_1$ norm between $I_{gt}$ and $I_{pred}$ that is weighed to contribute $\frac{1}{10}$ of the main reconstruction loss.  See examples in Fig. \ref{fig:losses}, first row.

\subsubsection*{Mask Loss $\mathcal{L}_{mask}$.} 
\begin{figure}[t]
\centering
\includegraphics[width=\linewidth]{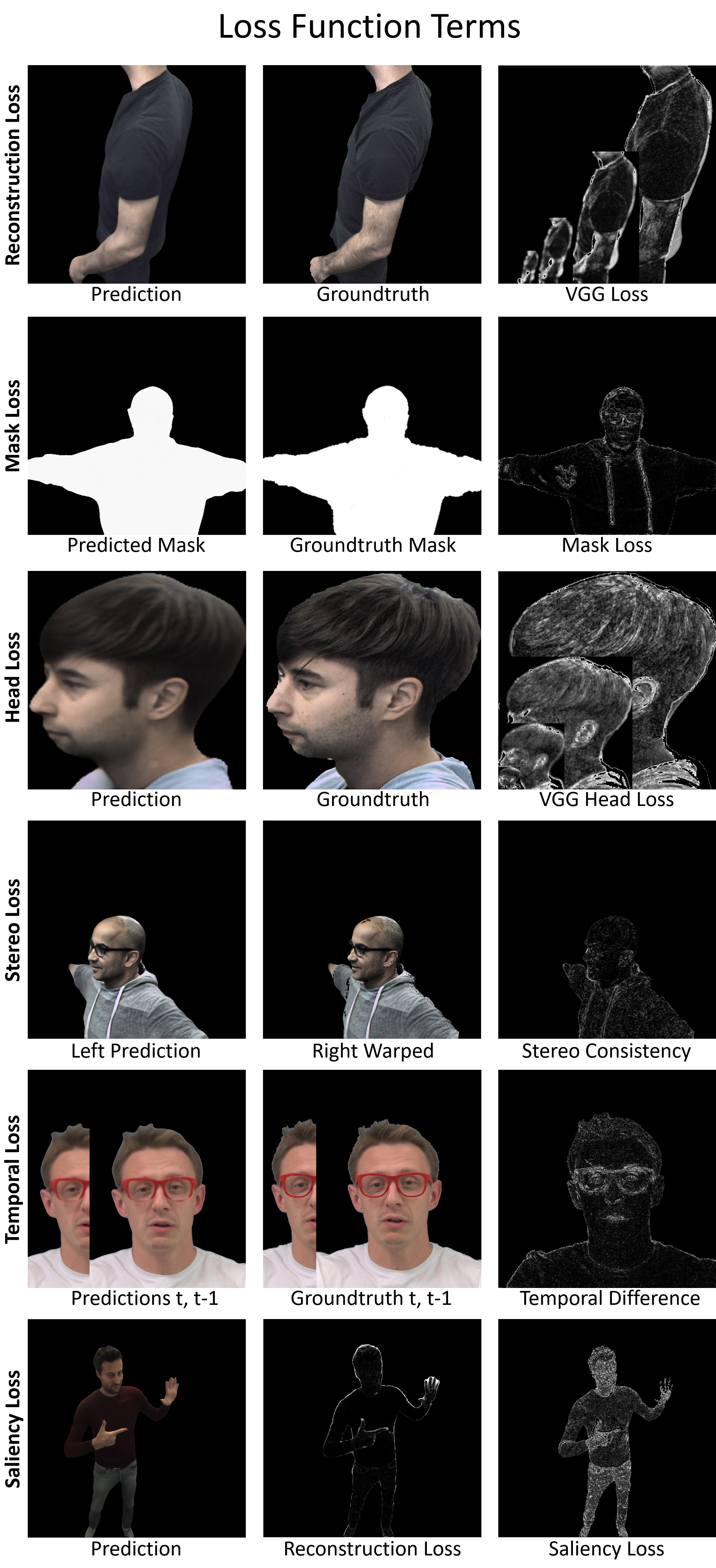}
\caption{Visualization of each term used in the loss function. See text for details.}
\label{fig:losses}
\end{figure}

The mask loss $\mathcal{L}_{mask}$ encourages the model to predict an accurate foreground mask $M_{pred}$.  This can be seen as a binary classification task. For foreground pixels we assign the value $y^+=1$, whereas for background pixels we use $y^-=0$. The final loss is defined as
\begin{equation}
\mathcal{L}_{mask} = \|M_{gt} - M_{pred}\|_{*}    
\end{equation}
where $\|\cdot\|_{*}$ is again the saliency re-weighted $\ell_1$ loss.  We also considered other classification losses such as a logistic loss but they all produced very similar results. An example of the mask loss is shown if Fig.  \ref{fig:losses}, second row.

\newcommand{\crop}{\mathcal{C}}
\subsubsection*{Head Loss $\mathcal{L}_{head}$.} The head loss focuses the network on the head to improve the overall sharpness of the face. Similar to the body loss, we use VGG16 to compute the loss in the feature space. In particular, we define  the crop $I^C$ for an image $I$ as a patch cropped around the head pixels as given by the segmentation labels of $I_{seg}$  and resized to $512 \times 512$ pixels. We then compute the loss as
\begin{equation}
\begin{split}
    \mathcal{L}_{head} = \sum_{i=1}^5 \Vert \VGG_i[M_{gt}^C \odot I_{gt}^C] - \VGG_i[M_{pred}^C \odot I_{pred}^C] \Vert_{*}~.
\end{split}
\end{equation}
For an illustration of the head loss, please see Fig.  \ref{fig:losses}, third row.

\subsubsection*{Temporal Loss $\mathcal{L}_{temporal}$.} {To minimize the amount of flickering between two consecutive frames, we design a temporal loss between a frame $I^t$ and $I^{t-1}$. A simple loss minimizing the difference between $I^t$ and $I^{t-1}$ would produce temporally blurred results, and thus we use a loss that tries to match the temporal gradient of the predicted sequence, i.e. $I_{pred}^t - I_{pred}^{t-1}$, with the temporal gradient of the ground truth sequence, i.e. $I_{gt}^t - I_{gt}^{t-1}$. In particular, the loss is computed as}
\begin{equation}
\begin{split}
    \mathcal{L}_{temporal} = \Vert ( I_{pred}^t -  I_{pred}^{t-1} ) - (I_{gt}^t -  I_{gt}^{t-1} ) \Vert_{1}~.
\end{split}
\end{equation}
Although recurrent architectures~\cite{rnn} have been proposed in the past to capture long range dependencies in temporal sequences, we found our non-recurrent architecture coupled with the temporal loss was able to produce temporally consistent outputs, with the added advantage of reduced inference time. {Another viable alternative consists of using optical flow methods to track correspondences between consecutive frames in the predicted images as well as in the groundtruth ones. The norm between these two motion fields can be used as a temporal loss. However this is bound to the quality of the flow method, and requires additional computation during the training. The proposed approach, instead, does not depend on perfect correspondences and works well for the purpose, i.e. to minimize the temporal flicker between frames.} Please see Fig.~\ref{fig:losses}, fifth row, for an example that illustrates the computed temporal loss.

\subsubsection*{Stereo Loss $\mathcal{L}_{stereo}$.} The stereo loss is specifically designed for VR and AR applications, when the network is applied on the left and right eye views. In this case, inconsistencies between both eyes might limit depth perception and result in discomfort for the user. One possible solution is to employ a second stereo ``witness'' camera placed at interpupillary distance with respect to the first one.

However, this might be unpractical due to bandwidth constraints. Therefore we propose an approach for those scenarios where such a stereo ground-truth is not available by proposing a loss that ensures self-supervised consistency in the output stereo images. 
\begin{figure*}[t]
\centering
\includegraphics[width=0.8\linewidth]{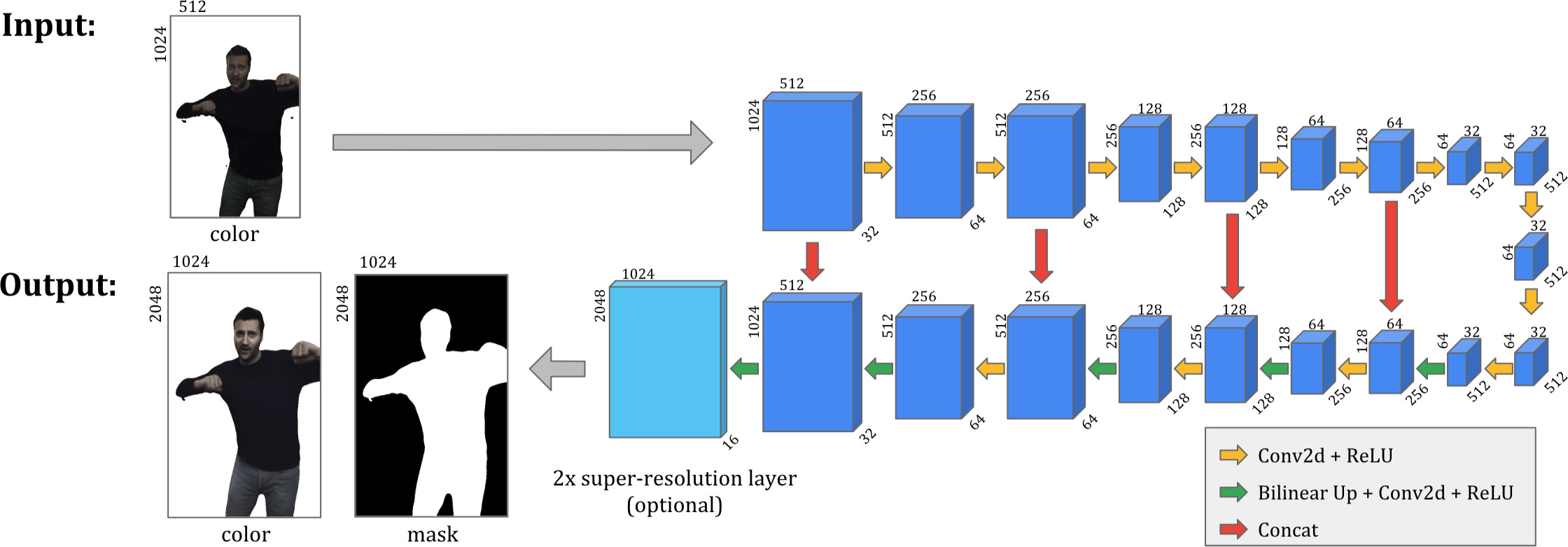}
\caption{LookinGood's fully convolutional deep architecture. We train the model for both left and right view that simulate a VR or AR headset. The architecture takes as input a low resolution image and produces a high quality rendering together with a foreground segmentation mask.}
\label{fig:architecture}
\end{figure*}

In particular, we render a stereo pair of the volumetric reconstruction and set each eye's image as input to the network, where the left image $I^L$ matches ground-truth camera viewpoint and the right image $I^R$ is rendered at $65$ mm along the x-coordinate. The right prediction $I_{pred}^R$ is then warped to the left viewpoint using the (known) geometry of the mesh and compared to the left prediction $I_{pred}^R$. We define a warp operator $I_{warp}$ using the Spatial Transformer Network (STN) \cite{stn}, which uses a bi-linear interpolation of $4$ pixels and fixed warp coordinates. We finally compute the loss as
\begin{equation}
\begin{split}
    \mathcal{L}_{stereo} = \Vert I_{pred}^L - I_{warp}(I_{pred}^R)\Vert_{1}~.
\end{split}
\end{equation}
Please see the fourth row of Fig  \ref{fig:losses} for examples that illustrate the stereo loss.

\newcommand{\sigmoid}{\rho}
\newcommand{\saliency}{\Upsilon}
\newcommand{\percentile}{\Gamma}
\newcommand{\residuals}{\mathbf{y}}
\subsubsection*{Saliency Re-weighing for Outlier Rejection.} The proposed losses receive a contribution from every pixel in the image (with the exception of the masked pixels). However, imperfections in the segmentation mask, may bias the network towards unimportant areas. Recently Lin~\etal\shortcite{focal} proposed to weigh pixels based on their difficulty: easy areas of an image are down-weighted, whereas hard pixels get higher importance. Conversely, we found pixels with the highest loss to be clear outliers, for instance next to the boundary of the segmentation mask, and they dominate the overall loss (see Fig.  \ref{fig:losses}, bottom row). Therefore, we wish to down-weight these outliers and discard them from the loss, while also down-weighing pixels that are easily reconstructed (e.g. smooth and textureless areas).
To do so, given a residual image $\mathbf{x}$ of size $W \times H \times C$, we set $\residuals$ as the per-pixel $\ell_1$ norm along channels of $\mathbf{x}$, and define minimum and maximum percentiles $p_{min}$ and $p_{max}$ over the values of $\residuals$.
% Those pixels that are above the $p_{max}$ percentile  do not contribute to the final loss, whereas those that are below $p_{min}$ percentile are quadratically down-weighted. 
We then define pixel's $p$ component of a ``saliency'' reweighing matrix of the residual $\residuals$ as
%\begin{equation}
%    \saliency_i(\mathcal{L} )= 
%    \begin{cases} 
%    \mathcal{L}_i \cdot \sigmoid(\mathcal{L}_i,p_{min}, \alpha_1), & \mbox{if }  \mathcal{L}_i \le p_{min}\\
%    \mathcal{L}_i, & \mbox{if } p_{min} < \mathcal{L} \le p_{max} \\ 
%    \mathcal{L}_i \cdot [1- \sigmoid(\mathcal{L}_i,p_{max}, \alpha_2)], & \mbox{if }  p_{max} < \mathcal{L} 
%    \end{cases}
%\end{equation} 

\begin{equation}
\saliency_p(\residuals)= 
     \begin{cases}
       1 &\quad\text{if }\residuals\in [\percentile(p_{min}, \residuals), \percentile(p_{max}, \residuals)] \\
       0 &\quad\text{otherwise.}
     \end{cases}
\end{equation}
% \begin{equation}
%     \saliency_p(\residuals)= \sigmoid[\residuals_,\percentile(p_{min}, \residuals), \alpha_1]\cdot[1- \sigmoid(\residuals_p,\percentile(p_{max}, \residuals), \alpha_2)] 
% \end{equation} 
% where $\sigmoid(x,c,\alpha) = \frac{1}{1-e^{-\alpha(x-c)}}$, 
where $\percentile(i, \residuals)$
 extracts the $i$'th percentile across the set of values in $\residuals$ and $p_{min}$, $p_{max}$, $\alpha_i$ are empirically chosen and depend on the task at hand (see Section \ref{sec:training}). We apply this saliency as a weight on each pixel of the residual $\residuals$ computed for $\mathcal{L}_{rec}$ and $\mathcal{L}_{head}$ as:
\begin{equation}
    \|\residuals\|_* =  \|\saliency(\residuals) \odot \residuals\|_1~.\label{eq:saliency-loss}
\end{equation}
where $\odot$ is the elementwise product.

Note that the we do not compute gradients with respect to the re-weighing function, and thus it does not need to be continuous for SGD to work. We experimented with a more complex, continuous formulation of $\saliency_p(\residuals)$ defined by the product of a sigmoid and an inverted sigmoid, and obtained similar results.

The effect of saliency reweighing is shown in the bottom row of Fig. \ref{fig:losses}.  Notice how the reconstruction error is along the boundary of the subject when no the saliency re-weighing is used. Conversely, the application of the proposed outlier removal technique forces the network to focus on reconstructing the actual subject. Finally, as byproduct of the saliency re-weighing we also predict a cleaner foreground mask, compared to the one obtained with the semantic segmentation algorithm used. Note that the saliency re-weighing scheme is only applied to the reconstruction, mask and head losses.

\subsection{Deep Architecture}
Our choice of the architecture is guided by two specific requirements: 1) the ability to perform inference in real-time 2) and effectiveness in the described scenario. Based on these requirements we resort to a U-NET like architecture~\cite{unet}. This model has shown impressive results in challenging novel viewpoint synthesis from 2D images problems~\cite{transfgrounded} and, moreover, can be run in real-time on high-end GPUs architectures. 

As opposed to the original system, we resort to a fully convolutional model (i.e. no max pooling operators).  Additionally, since it has been recently showed that deconvolutions can result in checkerboard artifacts~\cite{odena2016deconvolution}, we employed bilinear upsampling and convolutions instead. The overall framework is shown in Fig. \ref{fig:architecture}.

In more detail, our U-NET variation has a total of $18$ layers ($9$ encoding and $9$ decoding), with skip connections between the encoder and decoder blocks. The encoder begins with an initial $3 \times 3$ convolution with $N_{init}$ filters followed by a sequence of four ``downsampling blocks''. Each such block $i\in \{1, 2, 3, 4\}$ consists of two convolutional layers each with $N_i$ filters.  The first of these layers has a filter size $4 \times 4$, stride $2$ and padding $1$, whereas the second has a filter size of $3 \times 3 $ and stride $1$. Thus, each of the four block reduces the size of the input by a factor of $2$ due to the strided convolution.  Finally, two dimensionality preserving convolutions are performed (see far-right of Fig. \ref{fig:architecture}).  In all cases, the outputs of the convolutions are implicitly assumed to immediately pass through a ReLU activation function. Unless noted otherwise, we set $N_{init}=32$ and $N_i = G^i \cdot N_{init}$, where $G$ is the filter size growth factor after each downsampling block.

The decoder consists first of four ``upsampling blocks'' that mirror the ``downsampling blocks'' but in reverse.  Each such block $i \in \{4, 3, 2, 1\}$ consists of two convolutional layers.  The first layer bilinearly upsamples its input, performs a convolution with $N_i$ filters, and leverages a skip connection to concatenate the output with that of its mirrored encoding layer.  The second layer simply performs a convolution using $2N_i$ filters of size $3 \times 3$. Optionally, we add more upsampling blocks to produce images at a higher resolution than the input.

The final network output is produced by a final convolution with $4$ filters, whose output is, as per usual, passed through a ReLU activation function to produce the reconstructed image and a single channel binary mask of the foreground subject.

When our goal is to produce stereo images for VR and AR headsets, we simply run both left and right viewpoints using the same network (with shared weights). The final output is an improved stereo output pair.

\subsection{Training Details}
\label{sec:training}
% Superfluous paragraph.
% Our network architecture is implemented in Tensorflow.  It takes the mesh rendered at both stereo viewpoints as input and produces their respective outputs simultaneously. This is implemented using two branches of the network with shared weights that enables us to optimize the Tensorflow graph to produce outputs for VR rendering directly. 

We train the network using Adam \cite{adam} and weight decay \cite{tikhonov} until convergence (i.e. until the point where we no longer consistently observe drops in our losses). This was typically around 3 millions iterations for us. Training with Tensorflow on $16$ NVIDIA V100 GPUs with a batch size of 1 per GPU takes 55 hours.

% TODO(rohit): this does not stay, right?
% The network takes two (i.e. left and right view) $3$ channel RGB stereo patches of size $512 \times 512$ as input, and outputs two $4$ channel RGB plus mask pairs at an increased resolution of $1024 \times 1024$. Note that these images are crops from the original resolution of $1024 \times 512$ for the input and $2048 \times 1024$ for the output\footnote{Although our ground-truth cameras resolution is $4096 \times 2048$ this does not fit in memory during the training, therefore we downsample the images to $2048 \times 1024$.}. Crops are required only at training time due to the overhead needed to compute the loss functions; at test time we can fit in memory the full resolution images.

We use random crops for training, ranging from $512 \times 512$ to $960 \times 896$. Note that these images are crops from the original resolution of the input and output pairs. In particular, we force the random crop to contain the head pixels in $75\%$ of the samples, and for which we compute the head loss. Otherwise, we disable the head loss as the network might not see it completely in the input patch. This gives us the high quality results we seek for the face, while not ignoring other parts of the body as well. We find that using random crops along with standard $\ell$-2 regularization on the weights of the network is sufficient to prevent over-fitting. When high resolution witness cameras are employed the output is twice the input size.

% This paragraph is superfluous, we already say above that we set all the weights equal.
% The weights for each of our losses are chosen in a way that keeps their total numerical values comparable at beginning of training, and their values high enough to still have reasonable gradients towards the end of training. The weights $w_{mask}$,  $w_{temporal}$,  $w_{stereo}$ are set to $\frac{k_i}{W \times H}$ (i.e. they are down scaled by the size of the output image). The weight $w_{body}$ and $w_{head}$ are set to $\frac{k_j}{W \times H \times C}$ (i.e. they are down weighted by the width, height, and channels of the VGG activations). The constants $k_i$ are chosen in the range of $200$ to $500$ to prevent the losses from becoming too small near the end of training. These values do not perceptibly affect the final results, and we set to $k_i=300$ for all the losses.

The percentile ranges for the saliency re-weighing are empirically set to remove the contribution of the imperfect mask boundary and other outliers without affecting the result otherwise. We set $p_{max}=98$, and found that setting $p_{min}$ to values in $[25, 75]$ was acceptable, ultimately choosing $p_{min}=50$ for the reconstruction loss and $p_{min}=25$ for the head loss. We finally set both $\alpha_1 =\alpha_2 = 1.1$.

% !TEX root = main.tex
\section{Evaluation}
\begin{figure}[t]
\centering
\includegraphics[width=\columnwidth]{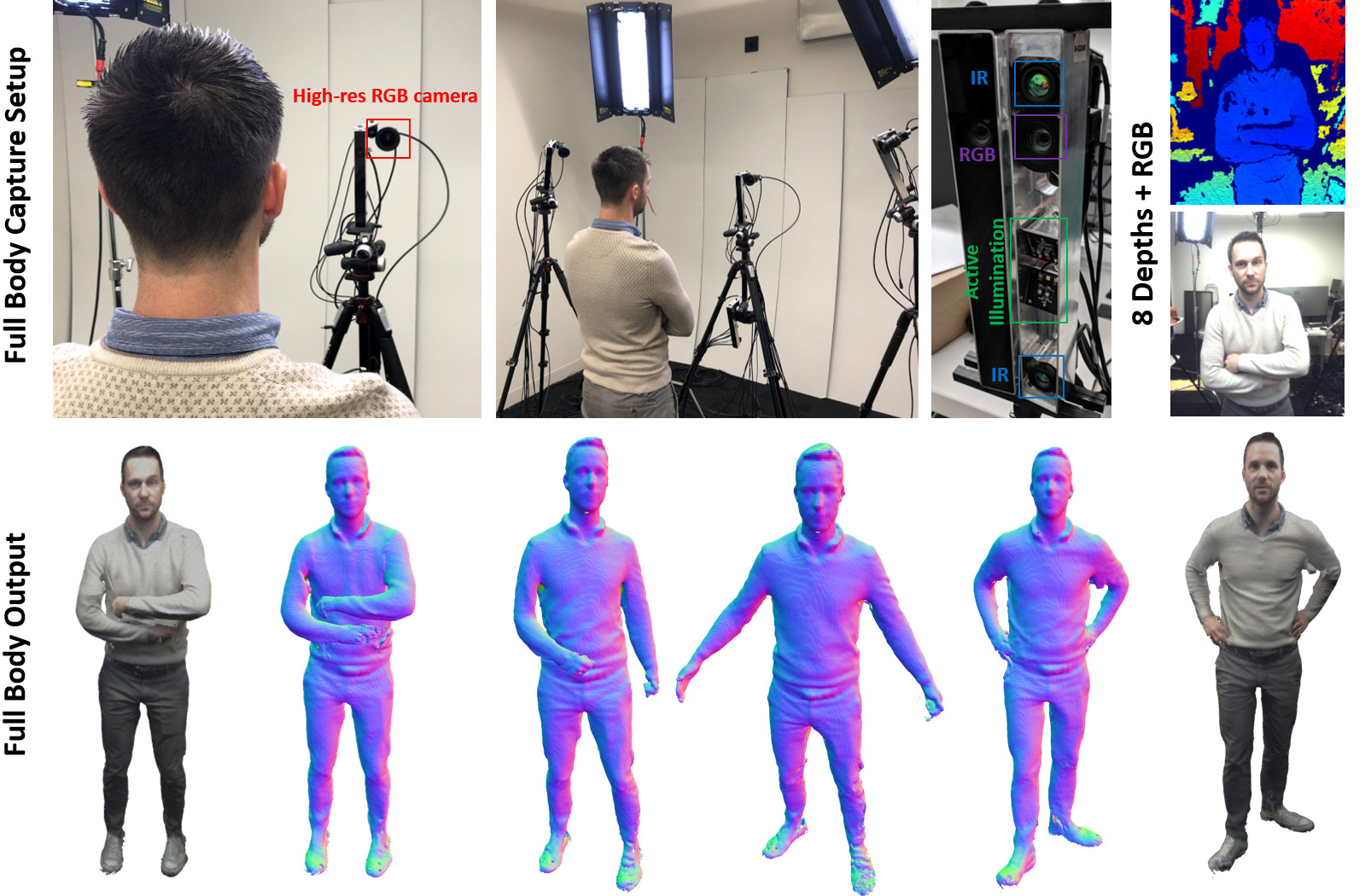}
\caption{Full body capture system. We implemented the method in \cite{dou17} where $8$ cameras are placed around the performer, who is reconstructed and tracked in real-time. We also added high resolution RGB cameras that are not used by the reconstruction system, but only at training time by the proposed machine learning method.}
\label{fig:fullbody_capture}
\end{figure}
\begin{table*}[t]
\caption{Quantitative evaluations on test sequences of subjects seen in training and subjects unseen in training. Photometric error is measured as the $\ell_1$-norm, and perceptual is the same loss based on VGG16 used for training. We fixed the architecture and we compared the proposed loss function with the same loss minus a specific loss term indicated in each columns. On seen subjects all the models perform similarly, whereas on new subjects the proposed loss has better generalization performances. Notice how the output of the volumetric reconstruction, i.e. the input to the network is  outperformed by all variants of the network.}
\begin{tabular}{| l | l | c | c | c | c | c | c| c |}
\hline
 & & Proposed & $-\mathcal{L}_{head}$& $-\mathcal{L}_{mask}$& -$Saliency$& -$\mathcal{L}_{stereo}$& $-\mathcal{L}_{temp}$ & Rendered Input \\ 
\hline
\textbf{Seen subjects} & Photometric Error & $0.0363$ & $0.0357$ & $0.0371$ & $0.0369$ & $\mathbf{0.0355}$ & $\mathbf{0.0355}$ & $0.0700$\\
 & PSNR & $\mathbf{29.2}$ & $\mathbf{29.2}$ & $28.2$ & $28.5$ & $29.0$ & $\mathbf{29.2}$ &  $25.0$\\
 & MS-SSIM & $0.956$ & $\mathbf{0.958}$ & $0.954$ & $0.954$ & $0.957$ & $0.957$ & $0.93$\\
 & Perceptual & $\mathbf{0.0658}$ & $0.121$ & $0.121$ & $0.103$ & $0.0963$ & $0.110$ & $0.1748$\\
\hline
\textbf{Unseen subjects} & Photometric Error & $\mathbf{0.0464}$ & $0.0498$ & $0.0506$ & $0.0510$ & $0.0465$ & $0.0504$ & $0.0783$\\
 & PSNR & $\mathbf{26.2}$ & $25.9$ & $25.5$ & $25.5$ & $26.0$ & $25.8$ & $24.05$ \\
 & MS-SSIM & $\mathbf{0.94}$ & $0.938$ & $0.929$ & $0.932$ & $0.937$ & $0.936$ & $0.9107$\\
 & Perceptual & $\mathbf{0.0795}$ & $0.168$ & $0.167$ & $0.136$ & $0.133$ & $0.157$ & $0.1996$\\
\hline
\end{tabular}
\label{tab:res}
\end{table*}

In this section we evaluate our system on two different datasets: one for single camera (upper body reconstruction) and one for multi view, full body capture. 

The \textbf{single camera dataset} comprises 42 participant of which 32 are used for training. For each participant, we captured four 10 second sequences, where they a) dictate a short text, with and without eyeglasses, b) look in all directions, and c) gesticulate extremely.
\begin{figure}[t]
\centering
\includegraphics[width=\linewidth]{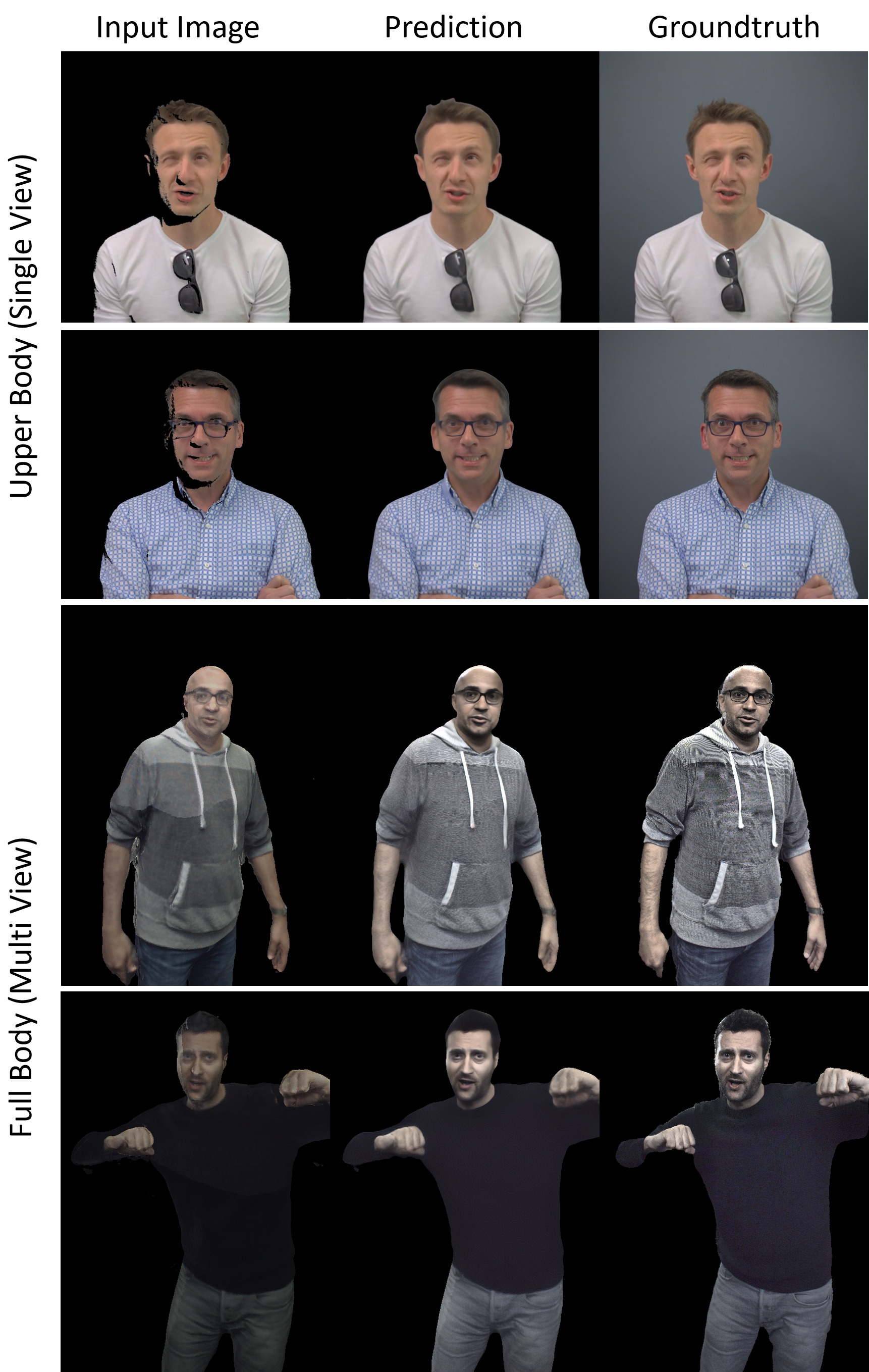}
\caption{Generalization on new sequences. We show here some results on known participant but unseen sequences. Notice how the method is able to in-paint missing areas correctly in the single camera case (top rows). Full body results show an improved quality and robustness to imprecision in the groundtruth mask (third row, right). The method also recovers from color and geometry inconsistencies (forth row, left).}
\label{fig:res_known}
\end{figure}

For the \textbf{full body capture data}, we recorded a diverse set of $20$ participants. Each performer was free to perform any arbitrary movement in the capture space (e.g. walking, jogging, dancing, etc.) while simultaneously performing facial movements and expressions. For each subject we recorded $10$ sequences of $500$ frames. 

We left $5$ subjects out from the training datasets to assess the performances of the algorithm on unseen people. Moreover, for some participants in the training set we left $1$ sequence (i.e. $500$ or $600$ frames) out for testing purposes.

\subsection{Volumetric Capture}
A core component of our framework is a volumetric capture system that can generate approximate textured geometry and render the result from any arbitrary viewpoint in real-time. For upper bodies, we leverage a high quality implementation of a standard rigid-fusion pipeline. For full bodies, we use a non-rigid fusion setup similar to Dou~\etal~\shortcite{dou17}, where multiple cameras provide a full $360^{\circ}$ coverage of the performer. 

\subsubsection*{Upper Body Capture (Single View).}
The upper body capture setting uses a single $1500 \times 1100$ active stereo camera paired with a $1600 \times 1200$ RGB view. To generate high quality geometry, we use a newly proposed method~\cite{espresso} that extends PatchMatch Stereo \cite{bleyer2011patchmatch} to spacetime matching, and produces depth images at 60Hz. We compute meshes by applying volumetric fusion~\cite{curless_and_levoy} and texture map the mesh with the color image as shown in Fig. \ref{fig:teaser} (top row).

In the upper body capture scenario, we mount a single camera at a $25^\circ$ degree angle to the side from where the subject is looking at at, of the same resolution as the capture camera. See Fig. \ref{fig:training_example}, top row, for an example of input/output pair.

\subsubsection*{Full Body Capture (Multi View).} For full body volumetric capture we implemented a system like the  Motion2Fusion framework \cite{dou17}. Following the original paper, we placed $16$ IR cameras and $8$ `low' resolution ($1280 \times 1024$) RGB cameras as to surround the user to be captured. The 16 IR cameras are built as 8 stereo pairs together with an active illuminator as to simplify the stereo matching problem (see Fig. \ref{fig:fullbody_capture} top right image for a breakdown of the hardware). We leverage fast, state of art disparity estimation algorithms~\cite{hyperdepth,fanello17_hashmatch,fanello2017ultrastereo,need4speed,sos} to estimate accurate depth. The non-rigid tracking pipeline follows the method of Dou~\etal\shortcite{dou17}. All the stages of the pipeline are performed in real-time. The output of the system consists of temporally consistent meshes and per-frame texture maps. In Fig. \ref{fig:fullbody_capture}, we show the overall capture system and some results obtained.

In the full body capture rig, we mounted $8$ `high' resolution ($4096 \times 2048$) witness cameras\footnote{Although our witness cameras resolution is $4096 \times 2048$ this does not fit in memory during the training, therefore we downsample the images to $2048 \times 1024$.} (see Fig. \ref{fig:fullbody_capture}, top left image). Examples of training examples are shown in Fig. \ref{fig:training_example}, bottom.

Note that both studied capture setups span a large number of use cases. The single-view capture rig does not allow for large viewpoint changes, but might be more practical, as it requires less processing and only needs to transmit a single RGBD stream, while the multi-view capture rig is limited to studio-type captures, but allows for complete free viewpoint video experiences.

\begin{figure*}[t]
\centering
\includegraphics[width=\linewidth]{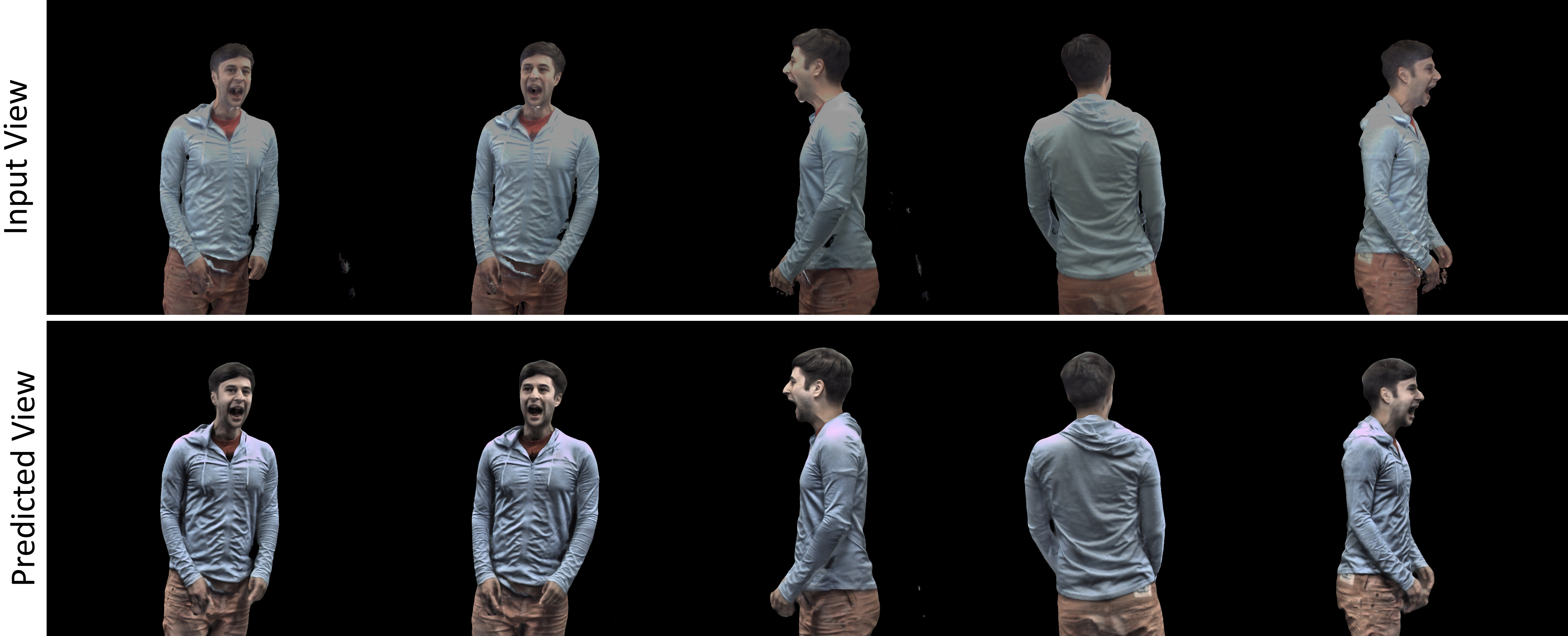}
\caption{Viewpoint Robustness. Notice how the neural re-rendering generalizes well w.r.t. to viewpoint changes, despite no training data was acquired for those particular views.}
\label{fig:viewpoint}
\end{figure*}
%% FIGURE
\begin{figure}[t]
\centering
\includegraphics[width=\linewidth]{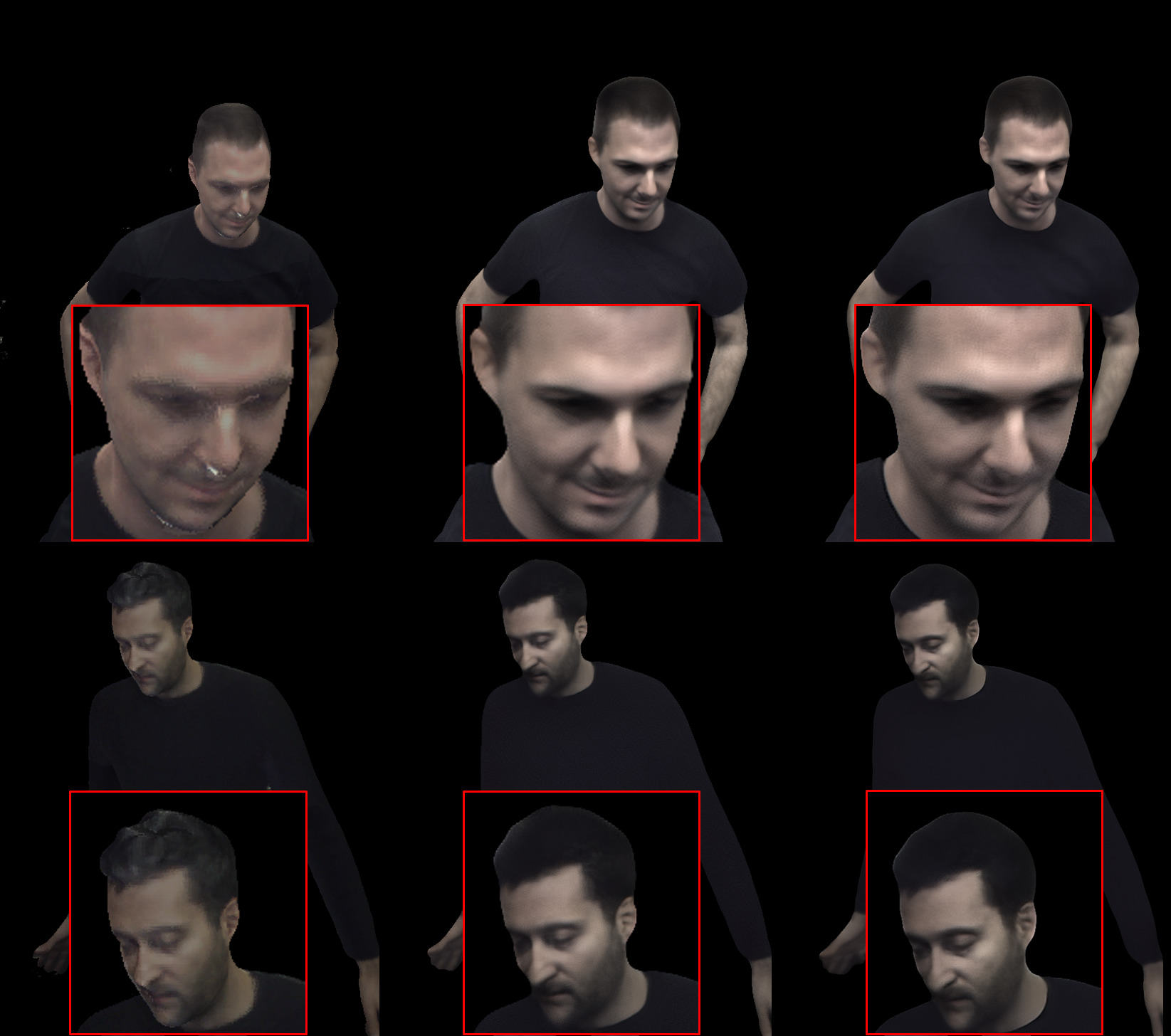}
\caption{Super-resolution experiment. The predicted output at the same resolution of the input shows more details. If we double the final resolution the final output is slightly sharper and it leads to better up-sampling especially around the edges.}
\label{fig:super_res}
\end{figure}

\subsubsection*{Experiments and Metrics.}
In the following, we test the performance of the system, analyzing the importance of each component. We perform two different analyses.  The first analysis is qualitative where we seek to assess the viewpoint robustness, generalization to different people, sequences and clothing. The second analysis is a quantitative evaluation on the proposed architectures. Since a real groundtruth metric is not available for the task, we rely on multiple perceptual measurements such as: PSNR, MultiScale-SSIM, Photometric Error, e.g. $\ell_1$-loss, and Perceptual Loss~\cite{perceptual}. Our experimental evaluation supports each design choice of the system and also shows the trade-offs between quality and model complexity. 

Many more results, comparisons and evaluations can be seen in the supplementary video (\href{http://youtu.be/Md3tdAKoLGU}{\texttt{http://youtu.be/Md3tdAKoLGU}}). Note that all results shown in the paper and in the supplementary video are on \emph{test sequences} that are not part of the training set.

\subsection{Qualitative Results}
Here we show qualitative results on different test sequences and under different conditions.

\subsubsection*{Upper Body Results (Single View).} In the single camera case, the network has to learn mostly to in-paint missing areas and fix missing fine geometry details such as eyeglasses frames. We show some results in Fig. \ref{fig:res_known}, top two rows. Notice how the method preserves the high quality details that are already in the input image and is able to in-paint plausible texture for those unseen regions. Further, thin structures such as the eyeglass frames get reconstructed in the network output. Note, that no super-resolution effects are observed, as the witness camera in the single view setup is of similar effective resolution than of the capture camera.

\subsubsection*{Full Body Results (Multi View).} The multi view case carries the additional complexity of blending together different images that may have different lighting conditions or have small calibration imprecisions. This affects the final rendering results as shown in Fig. \ref{fig:res_known}, bottom two rows. Notice how the input images have not only distorted geometry, but also color artifacts. Our system learns how to generate high quality renderings with reduced artifacts, while at the same time adjusting the color balance to the one of the witness cameras.

\subsubsection*{Viewpoint Robustness.} Although our groundtruth viewpoints are limited to a sparse set of cameras, in this section we demonstrate that the system is also robust to unseen camera poses. We implemented this by simulating a camera trajectory around the subject and show the results in Fig. \ref{fig:viewpoint}. More examples can be seen in the supplementary video.

\subsubsection*{Super-resolution.} Our model is able to produce more details compared to the input images. Results can be appreciated in Fig. \ref{fig:super_res}, where the predicted output at the same input resolution contains more subtle details like facial hair. Increasing the output resolution by a factor of $2$ leads to slightly sharper results and better up-sampling especially around the edges.

\begin{figure}[t]
\centering
\includegraphics[width=\linewidth]{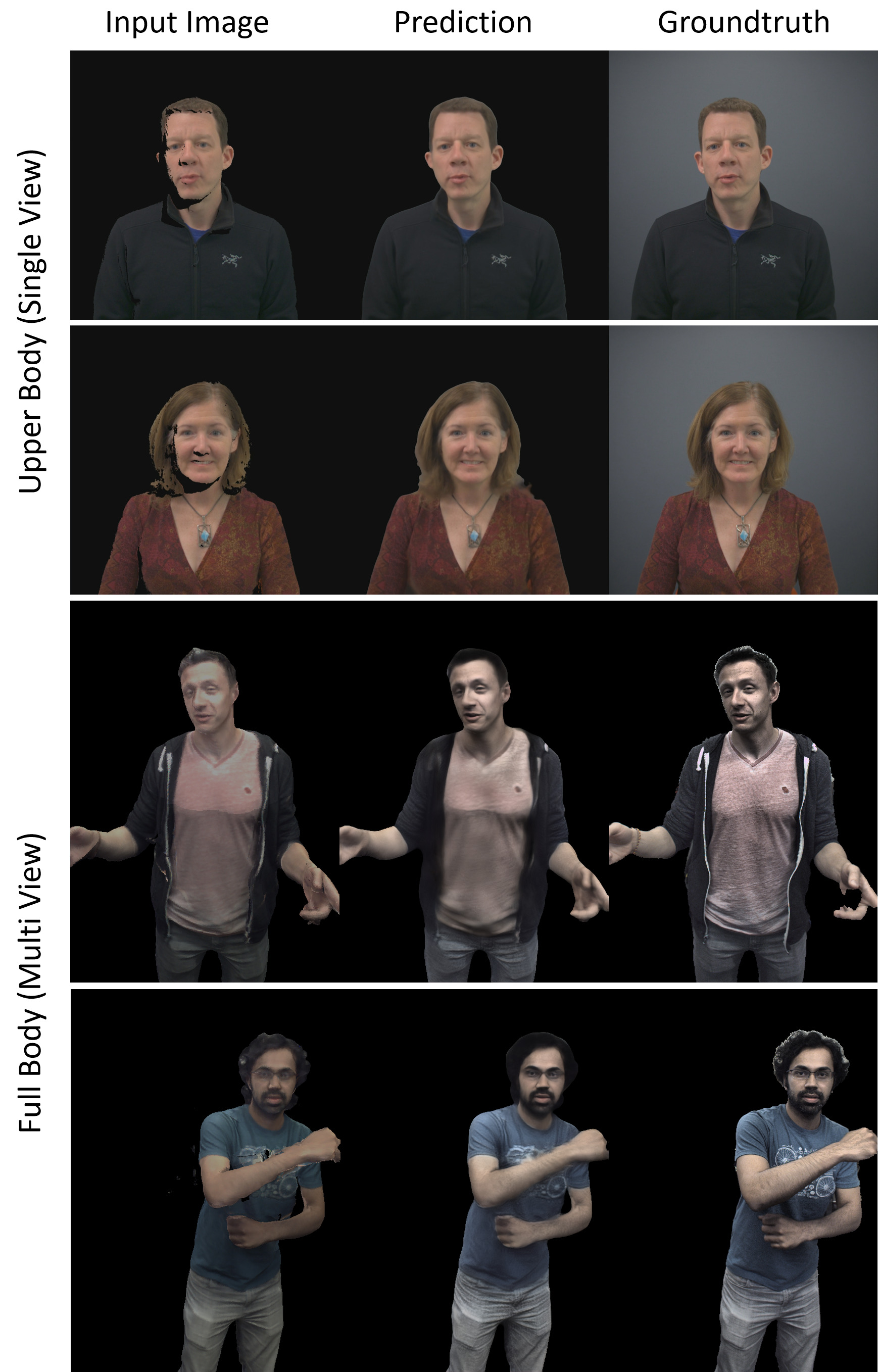}
\caption{Generalization on unseen subjects. The method correctly fills missing areas in the single camera case while maintaining high quality regions in the input. Full body results enhance the input and are robust to groundtruth mask outliers.}
\label{fig:res_unknown}
\end{figure}
\subsubsection*{Generalization: People, Clothing.} Generalization across different subjects are shown in Fig. \ref{fig:res_unknown}. For the single view case, we did not observe any substantial degradation in the results. For the full body case, although there is still a substantial improvement from the input image, the final results look less sharp. We believe that more diverse training data is needed to achieve better generalization performance on unseen participants.
\begin{figure}[t]
\centering
\includegraphics[width=\linewidth]{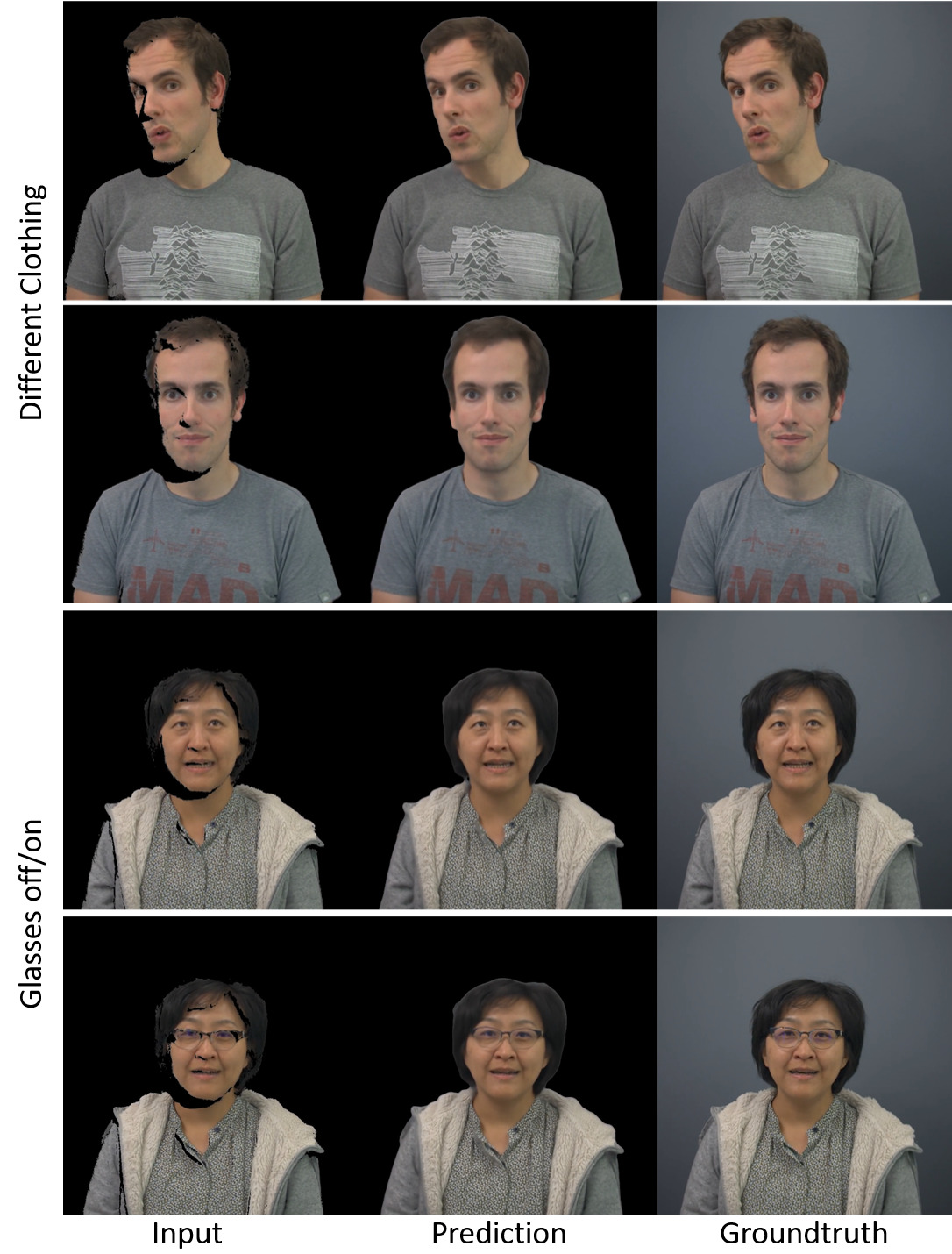}
\caption{The method performance is robust to changes in clothing (top) and eyewear (bottom).}
\label{fig:clothing}
\end{figure}

We also assessed the behavior of the system with different clothes or accessories. We show in Fig. \ref{fig:clothing} examples of such situations: a subject wearing different clothes, and another with and without eyeglasses. The system correctly recovers most of the eyeglasses frame structure even though they are barely reconstructed by the traditional geometrical approach due to their fine structures.

\begin{figure}[t]
\centering
\includegraphics[width=\linewidth]{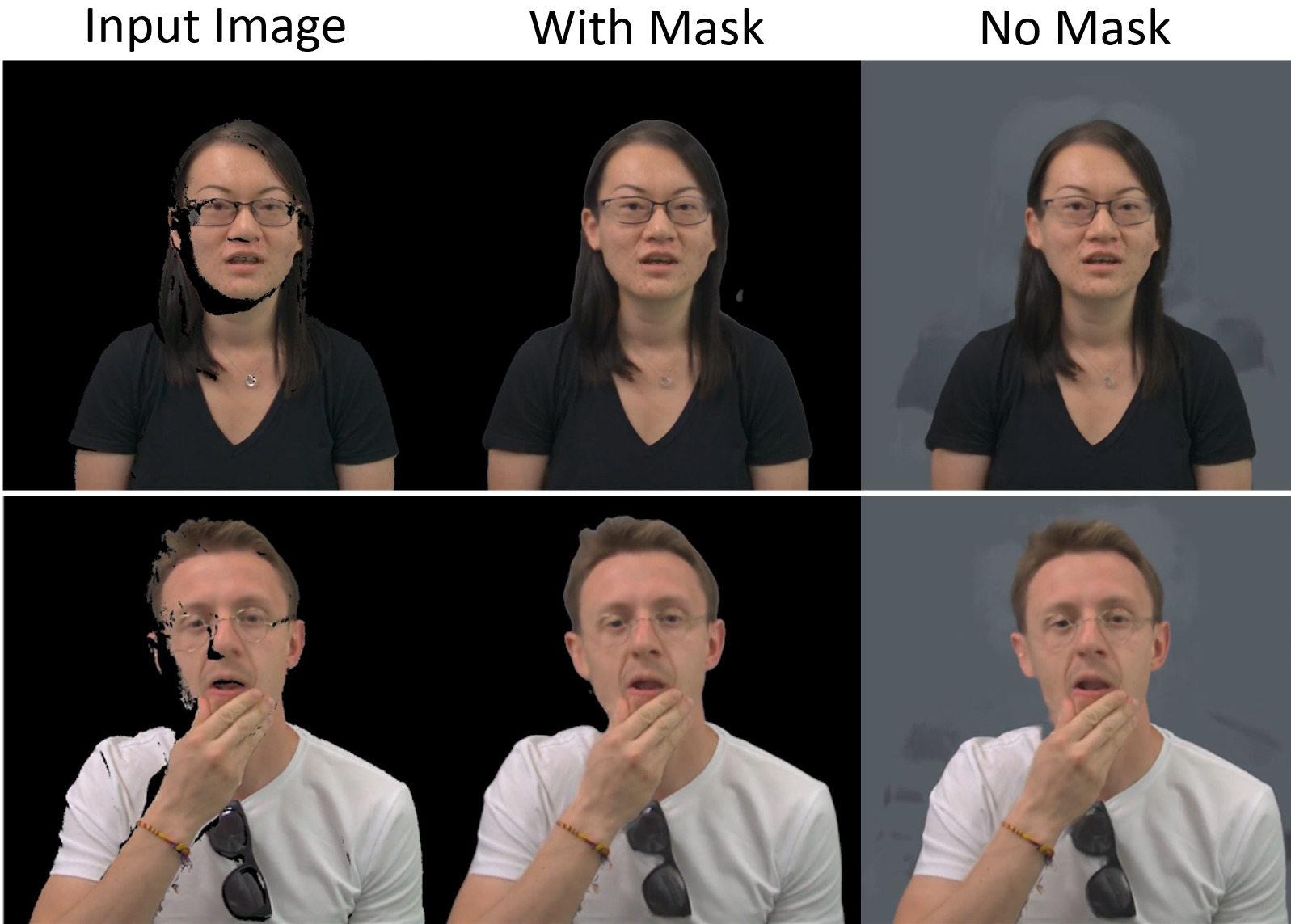}
\caption{Effect of the predicted foreground mask. Notice how when no mask is used the network tries to predict a noisy version of the background.}
\label{fig:mask_qual}
\end{figure}

\subsection{Ablation Study}
We now show the importance of the different components of the method. The main quantitative results are summarized in Table \ref{tab:res}, where we computed multiple statistics for the proposed model and all its variants. In the following we comment on the findings.

\subsubsection*{Segmentation Mask.} The segmentation mask plays an important role in in-painting missing parts, discarding the background and preserving input regions. As shown in Fig. \ref{fig:mask_qual}, the model without the foreground mask hallucinates parts of the background and does not correctly follow the silhouette of the subject. This behavior is also confirmed in the quantitative results in Table \ref{tab:res}, where the model without the $\mathcal{L}_{mask}$ performs worse compared to the proposed model.

\begin{figure}[t]
\centering
\includegraphics[width=\linewidth]{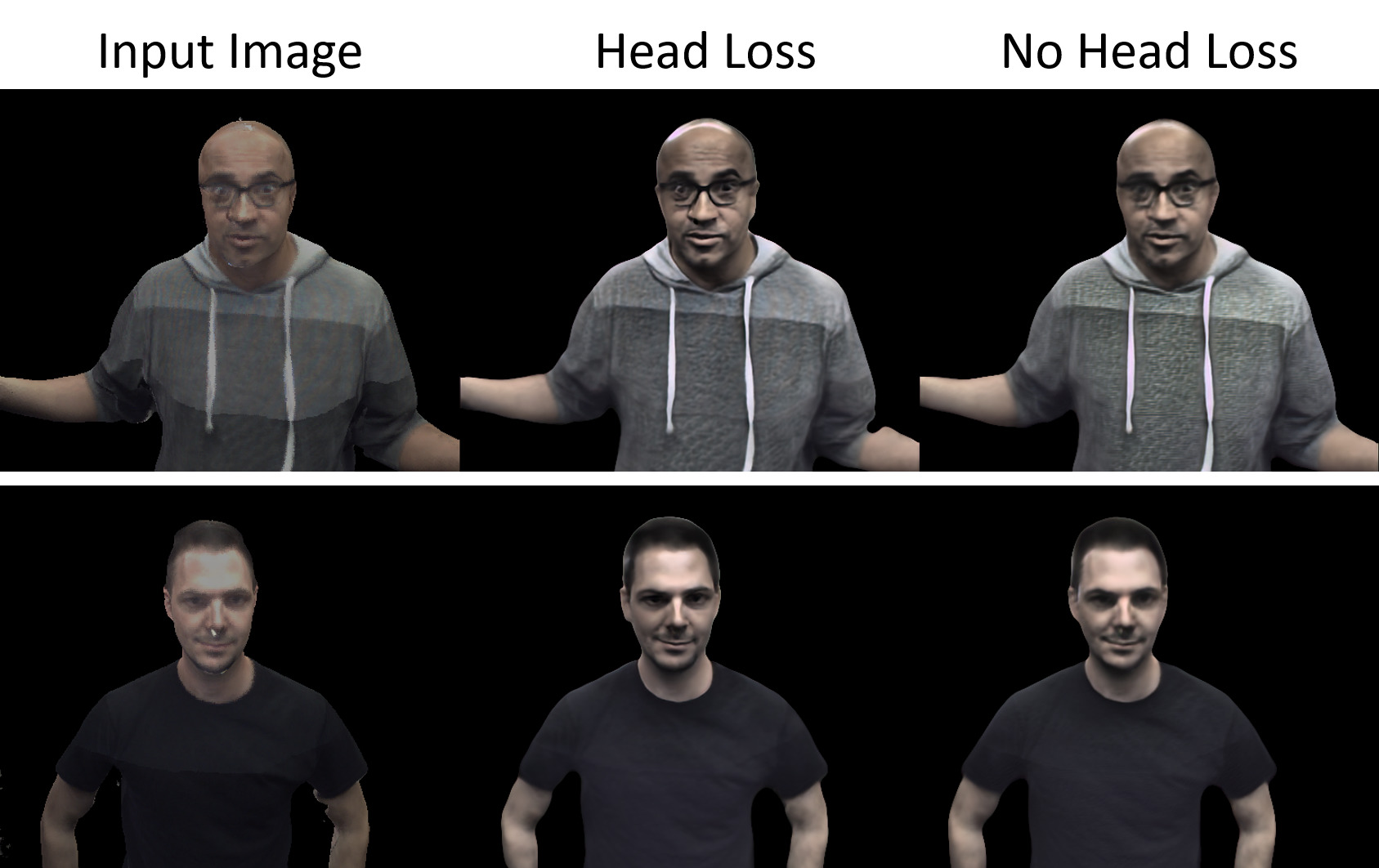}
\caption{Effect of the proposed head loss $\mathcal{L}_{head}$. Notice how the predicted output is sharper when the head loss is used. Best viewed in the digital version.}
\label{fig:head_loss_qual}
\end{figure}

\subsubsection*{Head Loss.} The loss on the cropped head regions encourages sharper results on faces. Previous studies \cite{holoportation} found that artifacts in the face region are more likely to disturb the viewer. We found the proposed loss to greatly improve this region. Although the numbers in Table~\ref{tab:res} are comparable, there is a huge visual gap between the two losses, as shown in Fig.~\ref{fig:head_loss_qual}. Notice how without head loss the results are oversmoothed and facial details are lost. Whereas the proposed loss not only upgrades the quality of the input, but it also recovers unseen features. 

\subsubsection*{Temporal and Stereo Consistency.} 
\begin{figure}[t]
\centering
\includegraphics[width=\linewidth]{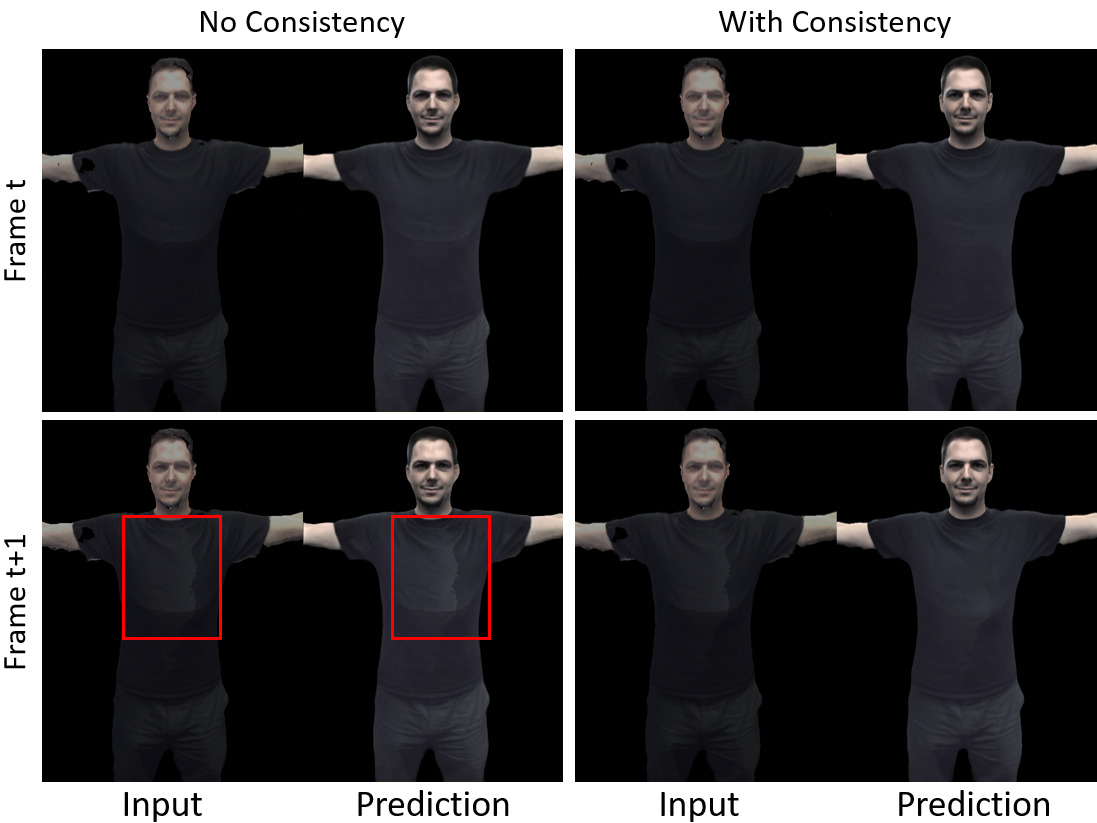}
\caption{
Effects of temporal and stereo consistency: two consecutive frames where the input flickers due to a texture atlas change. On the left, a model trained without consistency losses exhibits the input temporal inconsistency shown in the highlighted region. However, this is significantly reduced in the model trained with consistency losses.}
\label{fig:consistency}
\end{figure}

Stable results across multiple viewpoints have already been shown in Fig. \ref{fig:viewpoint}. The metrics in Table~\ref{tab:res} show that removing temporal and stereo consistency from the optimization sometimes may outperform the model trained with the full loss function. However, this is somehow expected since the metrics used do not take into account important factors such as temporal and spatial flickering. The effects of the temporal and stereo loss are visualized in Fig.~\ref{fig:consistency}.

\subsubsection*{Saliency Reweighing.} 
\begin{figure}[t]
\centering
\includegraphics[width=\columnwidth]{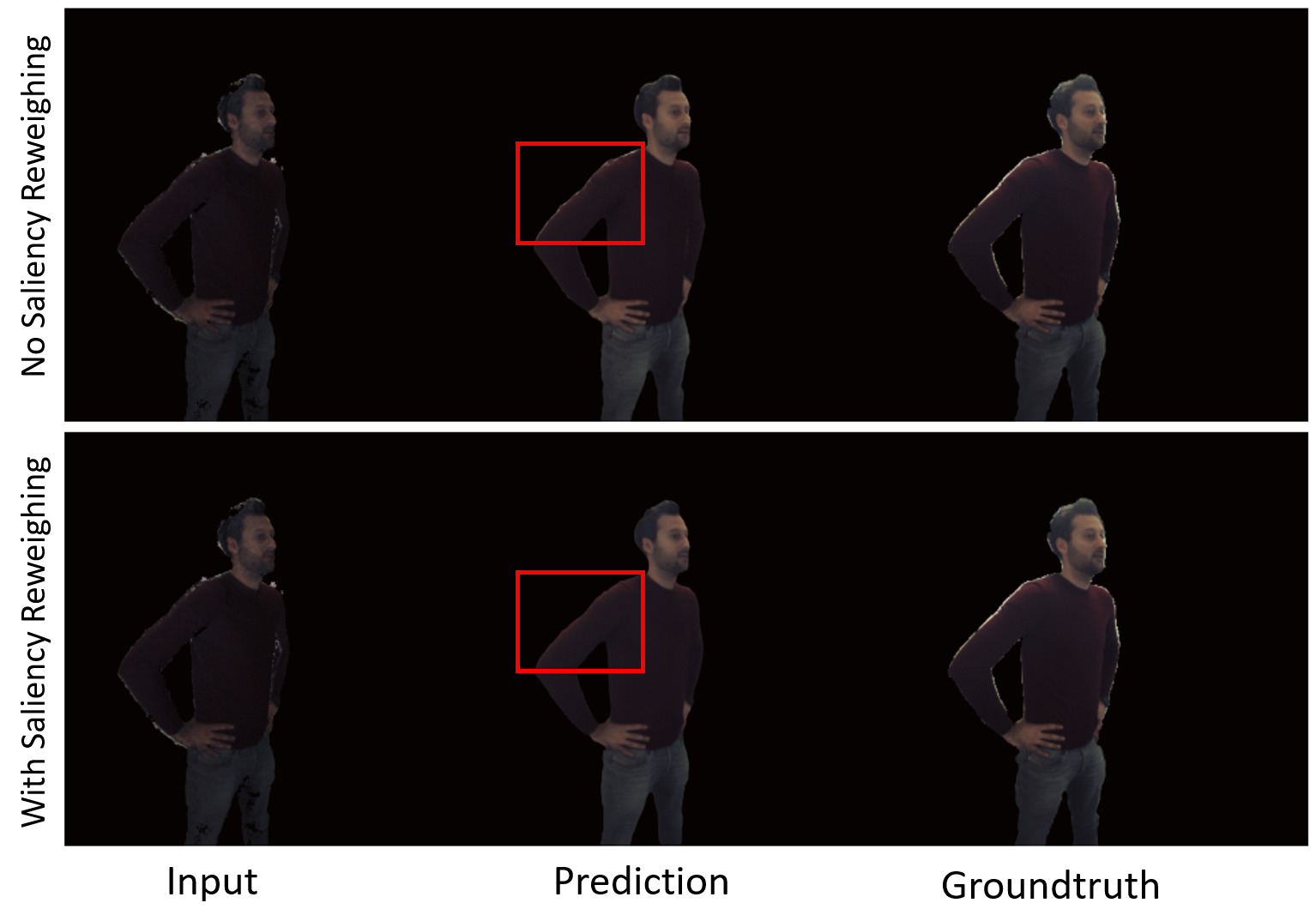}
\caption{
The proposed saliency reweighing scheme of the losses in the bottom reduces the influence of the mask outliers around the silhouette of the subject in the bottom row, while the model trained without reweighing displays white artifacts in silhouette due to outliers in the segmentation mask (top). Best seen in the digital version of the paper.}
\label{fig:saliency}
\end{figure}

The saliency reweighing reduces the effect of outliers as shown in Fig.~\ref{fig:losses}. This can also be appreciated in all the metrics in Table~\ref{tab:res}: indeed the models trained without the saliency reweighing perform consistently worse. Figure \ref{fig:saliency} shows how the model trained with the saliency reweighing is more robust to outliers in the groundtruth mask.

\subsubsection*{Model Size.} 
\begin{figure}[t]
\centering
\includegraphics[width=\columnwidth]{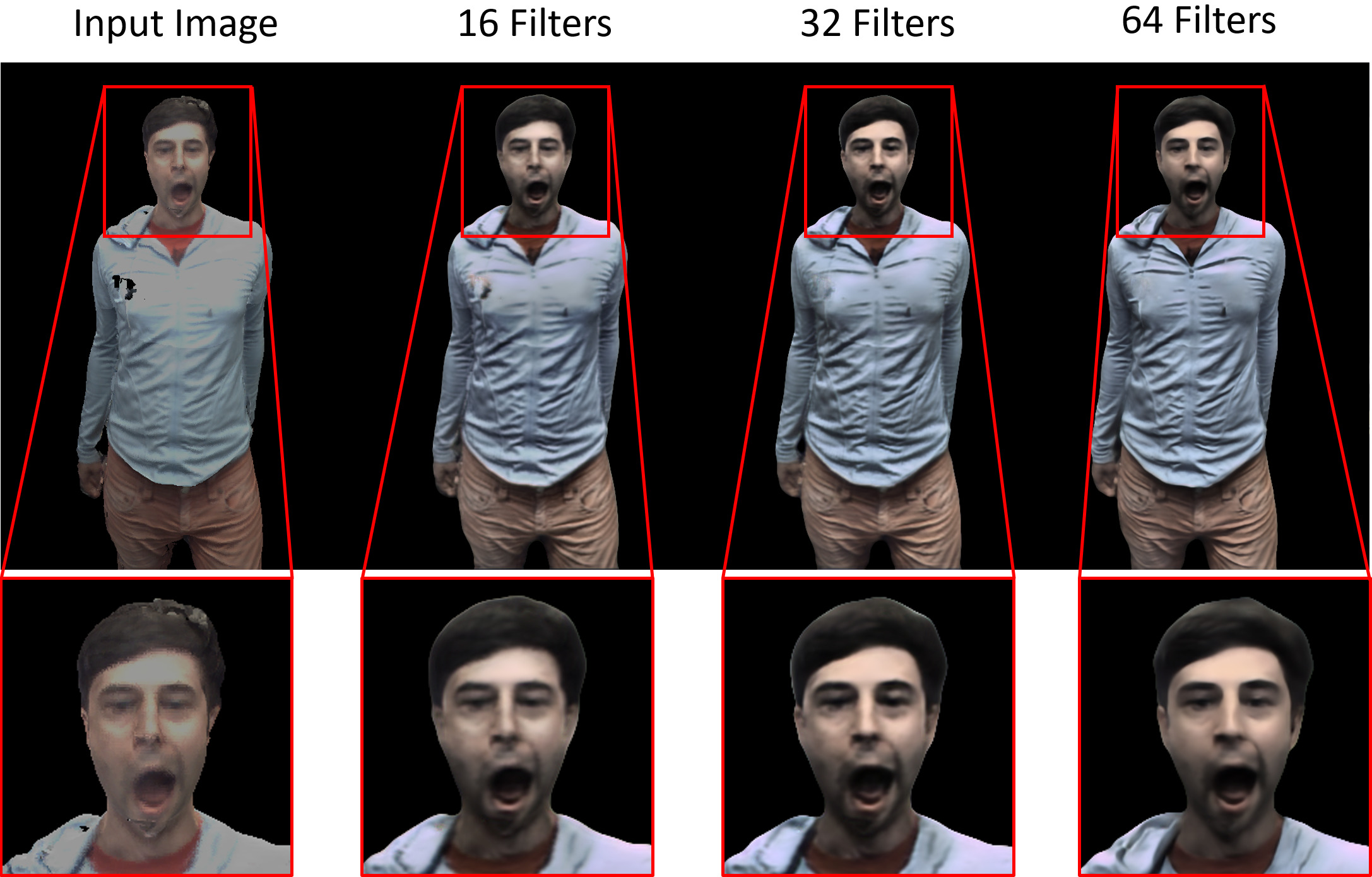}
\caption{
Model complexity analysis. The model starting with $N_{init}=64$ filters in the first layer leads to the sharpest results.}
\label{fig:model_size}
\end{figure}

We also assess the importance of the model size. We trained three different networks, starting with $N_{init}=16,32,64$ filters respectively. In Fig.~\ref{fig:model_size} we show qualitative examples of the three different model. As expected, the biggest network achieves the better and sharper results on this task, showing that the capacity of the other two architectures is limited for this problem. 
%However, in later experiments, we observed that adding a GAN loss can reduce the quality difference between the different sized networks.

% !TEX root = main.tex
%\section{User Study}

%User Study

%Show videos (or demo) with beautification on/off

%Ask which one the user prefers

%Ask which one is closer to a given GT image

%Ask which one is more photo-realistic

%Comparison with point cloud rendering
\begin{figure}[t]
\centering
\includegraphics[width=\linewidth]{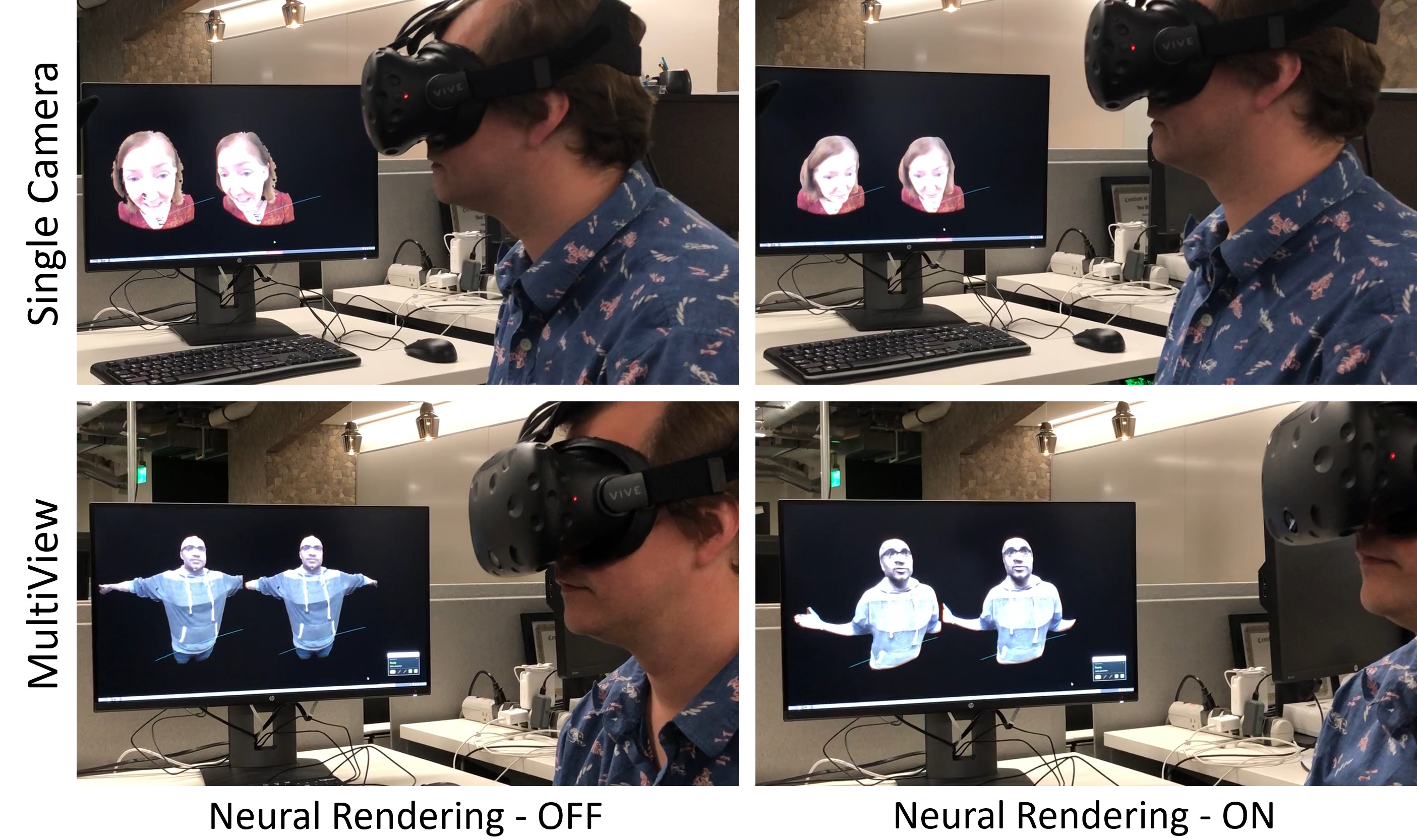}
\caption{Real-time demo showing neural re-rendering on a single camera reconstruction (top) and full body volumetric reconstruction (bottom).}
\label{fig:demo}
\end{figure}
\section{Real-time Free Viewpoint Neural Re-Rendering}
We implemented a real-time demonstration of the system, as shown in Fig. \ref{fig:demo}. The scenario consists of a user wearing a VR headset watching volumetric reconstructions. We render left and right views with the head pose given by the headset and feed them as input to the network. The network generates the enhanced re-renderings that are then shown in the headset display.

Latency is an important factor when dealing with real-time experiences. Instead of running the neural re-rendering sequentially with the actual display update, we implemented a late stage reprojection phase \cite{van2016asynchronous,evangelakos2016extended}. In particular, we keep the computational stream of the network decoupled from the actual rendering, and use the current head pose to warp the final images accordingly.
\begin{figure}[t]
\centering
\includegraphics[width=\linewidth]{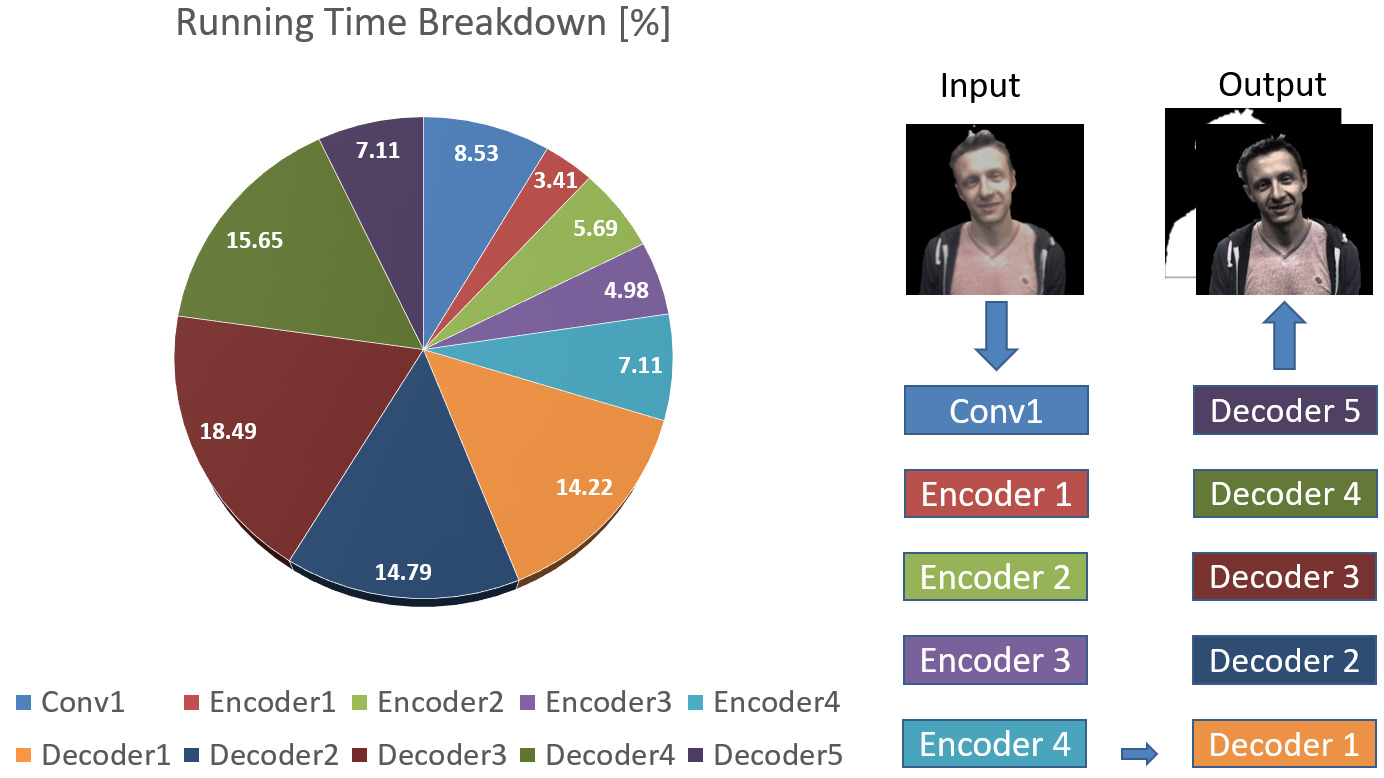}
\caption{Running time breakdown in percentage of the current model. Notice how most of the time is spent at the decoding stages due to the skip connections. }
\label{fig:timepie}
\end{figure}
\subsection{Neural Re-Rendering Runtime}
{We assessed the run-time of the system  using a single NVIDIA Titan V. We considered the model with $N_{init}=32$ filters where input and output are generated at the same resolution ($512 \times 1024$). Using the standard TensorFlow graph export tool, the average running time to produce a stereo pair with our neural re-rendering is around $92$ms, which is not sufficient for real-time applications. Therefore we leveraged NVIDIA TensorRT, which performs inference optimization for a given deep architecture. Thanks to this tool, a standard export with $32$bits floating point weight brings the computational time down to $47$ms. Finally, we exploited the optimizations implemented on the NVIDIA Titan V, and quantize the network weights using 16-bit floating point. This allows us reaching the final run-time of $29$ms per stereo pair, with no loss in accuracy, hitting the real-time requirements.}

{We also profiled each block of the network to find potential bottlenecks. We report the analysis in Fig.~\ref{fig:timepie}. The encoder phase needs less than $40\%$ of the total computational resources. As expected, most of the time is spent in the decoder layers, where the skip connections (i.e. the concatenation of encoder features with the matched decoder), leads to large convolution kernels. Possible future work consists of replacing the concatenation of the skip connections with sum, which would reduce the features size.}

\subsection{User Study}
{We performed a small qualitative user study on the results of the output system, following an approach similar to~\cite{visual_turing_test}. We recruited $10$ subjects and prepared $12$ short video sequences showing the renderings of the capture system, the predicted results and the target witness views (masked with the semantic segmentation as described in Section \ref{sec:imen}). The order of the videos was randomized and we selected sequences containing both seen subjects and unseen subjects.}

{We asked the participants whether they preferred the renders of the performance capture system (i.e. the  input to our algorithm), the re-rendered versions using neural re-rendering, or the masked ground truth image, i.e. $M_{gt} \odot I_{gt}$. Not surprisingly, $100\%$ of the users agreed that the output of the neural re-rendering was better compared to the renderings from the volumetric capture systems. Also, the users did not seem to notice substantial differences between seen and unseen subjects. Unexpectedly, $65\%$ of the subjects preferred the output of our system even compared to the groundtruth: indeed the participants found the predicted masks using our network to be more stable than the groundtruth masks used for training, which suffers from more inconsistent predictions between consecutive frames. However all the subjects agreed that groundtruth is still sharper, therefore higher resolution than the neural re-rendering output, and more must be done in this direction to improve the overall quality.}

% !TEX root = main.tex
\section{Discussion, Limitations and Future Work}
\begin{figure}[t]
\centering
\includegraphics[width=\columnwidth]{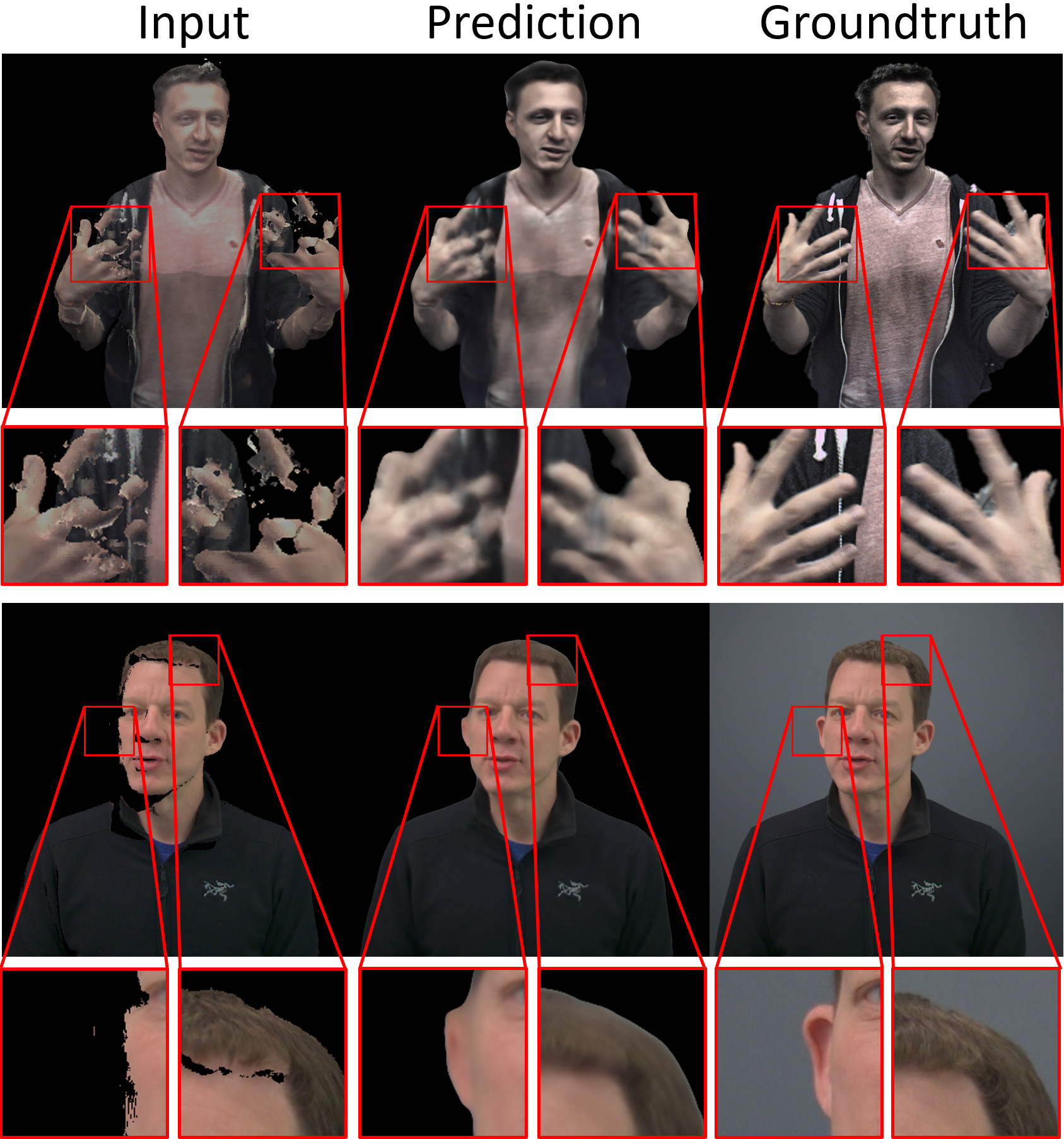}
\caption{
Limitations of our system. Top: when the input is particularly corrupted in both geometry and texture, the produced output is blurry. Bottom: hallucinated parts may not resemble the real image, as shown in the highlighted region on the left. As well, the neural re-rendering may lose some high frequency details present in the input in regions without artifacts, due to the limited model size, as shown in the highlighted hair region.}
\label{fig:limitations}
\end{figure}
We presented ``\emph{LookinGood}'', the first system that uses machine learning to enhance volumetric videos in real-time. We carefully combined geometric non-rigid reconstruction pipelines, such as \cite{dou17}, with recent advances in deep learning, to produce higher quality outputs. We designed our system to focus on people's faces, discarding non-relevant information such as the background. We proposed a simple and effective solution to produce temporally stable renderings and devoted particular attention to VR and AR applications, where left and right views must be consistent for an optimal user experience.

We found the main limitation of the system to be the lack of training data. Indeed, whereas unseen sequences of known subjects still produce very high quality results, we noticed a graceful degradation of the quality when the participant was not in the dataset (see Fig. \ref{fig:res_unknown}). When the input is very partially corrupted, the model hallucinates blurry results, as shown in Fig. \ref{fig:limitations}, top row. In addition, missing parts are sometimes oversmoothed. Although a viable solution consists of acquiring more training examples, we prefer to focus our future efforts on more intelligent deep architectures. We will, for instance, reduce the capture infrastructure by leveraging recent deep architectures for accurate geometry estimation \cite{stereonet,activestereonet}; furthermore, we will introduce a calibration phase where a new user will be able to quickly personalize the system for better run-time performance and accuracy. Finally, by leveraging semantic information, such as pose estimation and tracking \cite{totalcapture}, we will make the problem even more tractable when multi-view rigs are not available.

\begin{acks}
We thank Jason Lawrence, Harris Nover, and Supreeth Achar for continuous feedback and support regarding this work.
\end{acks}

\bibliographystyle{ACM-Reference-Format}
\bibliography{references}
\end{document}